\documentclass[10pt,journal]{IEEEtran}
\newtheorem{myDef}{Definition}

\usepackage{graphicx} 
\usepackage[switch]{lineno}
\ifCLASSOPTIONcompsoc
  \usepackage[nocompress]{cite}
\else
  \usepackage{cite}
\fi
\ifCLASSINFOpdf
\else
\fi
\begin{document}
%
\title{Generalized Constraints as A New Mathematical Problem in Artificial Intelligence: A Review and Perspective}
%
%
%
%

\author{Bao-Gang~Hu,~\IEEEmembership{Senior Member,~IEEE,}
        and~Han-Bing~Qu,~\IEEEmembership{Member,~IEEE,} 
\thanks{Manuscript created September 19, 2020.(Corresponding Author: Bao-Gang Hu)}
\thanks{B.-G. Hu is with National Laboratory of Pattern Recognition, Institute of Automation, Chinese Academy of Science, and University of Chinese Academy of Sciences, Beijing, 100091, China. }
\thanks{H.-B. Qu is with Beijing Institute of New Technology Applications, Beijing Academy of Science and Technology, Beijing, 100094, China. }
}

\maketitle.

\begin{abstract}
In this comprehensive review, we describe a new mathematical problem in artificial intelligence (\textbf{AI}) from a mathematical modeling perspective, following the philosophy stated by Rudolf E. Kalman that {\it ``Once you get the physics right, the rest is mathematics"}. The new problem is called ``{\it Generalized Constraints} (\textbf{GCs})'', and we adopt GCs as a general term to describe any type of prior information in modelings. To understand better about GCs to be a general problem, we compare them with the {\it conventional constraints} (\textbf{CCs}) and list their extra challenges over CCs. In the construction of AI machines, we basically encounter more often GCs for modeling, rather than CCs with {\it well-defined} forms. Furthermore, we discuss the ultimate goals of AI and redefine  {\it transparent},  {\it interpretable}, and  {\it explainable AI} in terms of comprehension levels about machines. We review the studies in relation to the GC problems although most of them do not take the notion of GCs. We demonstrate that if AI machines are simplified by a coupling with both {\it knowledge-driven} submodel and {\it data-driven} submodel, GCs will play a critical role in a knowledge-driven submodel as well as in the coupling form between the two submodels. Examples are given to show that the studies in view of a generalized constraint problem will help us perceive and explore novel subjects in AI, or even in mathematics, such as {\it generalized constraint learning} (\textbf{GCL}).
\end{abstract}




\section{Introduction}
\label{sec:introduction}

\IEEEPARstart{O}{nce} you get the physics right, the rest is mathematics \cite{Kalman:08:LEC}. This statement by Kalman is particularly true to the study of artificial intelligence (\textbf{AI}). In contrast to the natural intelligence displayed by humans or other lives, AI demonstrates its intelligence by machines (or tools, systems, models in other terms) programmed from a computer language. The statement will direct us to seek the fundamental of AI at a mathematical level rather than to stay at application levels. Therefore, when deep learning (\textbf{DL}) advanced AI to a new wave, one question seems to be: \textbf{"Do we encounter any new mathematical, yet general, problem in AI"}?

 In this paper, we take the notion of ``{\it Generalized Constraints} (\textbf{GCs})'' and consider it as a new mathematical problem in AI. The related backgrounds are given below.
   
\subsection{Three core terms} 
 
  \begin{myDef} Conventional constraint (\textbf{CC}) refers to a constraint in which its representation is fully known and given in a structured form. 
 \end{myDef}
  \begin{myDef} Prior information (\textbf{PI}) is any information or knowledge about the particular things (such as problem, data, fact, etc.) that is known for someone. 
 \end{myDef}
 \begin{myDef} Generalized constraint (\textbf{GC}) is a term used in mathematical modelings to describe any related prior information. 
 \end{myDef}

 From the definitions above, we can describe their relations by using a set notation: 
 \begin{equation}\label{equ_1}
 \mbox{PI}  = \mbox{GC} \supset \mbox{CC}, 
 \end{equation}
 where CC is a subset of GC, and GC is equal to PI. When the term PI appears in daily life, the term GC stresses a mathematical meaning in  modeling. 
 For simplifying discussions, we take PI as a general term which may be called {\it prior knowledge}, 
 {\it prior fact}, {\it specification}, {\it bias}, {\it hint}, {\it context}, {\it side information},  
 {\it invariance}, {\it idea}, {\it hypothesis}, {\it principle}, {\it theory}, {\it common sense}, etc. 
 In \cite{Hu:09:IS}, Hu et al. pointed out that PI usually exhibits one or a combination of features in modelings, such as incomplete information and unstructured form in its representation. They showed several examples about incomplete information in GCs, such as, 
 $1-ax^2- by^2 > 0$, 
 where $a$ and $b$ are unknown parameters.  
 
\subsection{Extra challenges of GCs over CCs}
 
Duda et al. pointed out that \cite{Duda:01:Book}: ``\textit{incorporating prior knowledge can be far more subtle and difficult}''. For a better understanding about CCs and GCs, we present an overall comparison between them in respect to several aspects (Table \ref{Table1}). One can see that GCs do not only enlarge the application domains and the representation forms over CCs, but also add a significant amount of extra challenges in AI studies. For example, the new subjects may appear from the study of GCs. Some of GCs may involve a transformation between linguistic prior and computational representation, which is a difficult task because one may face a ``{\it semantic gap}'' \cite{Hu:15:CIAC}. We will discuss those extra challenges further in the later sections. 
\begin{table*}[ht]
\caption{Comparisons between CCs and GCs}
\centering
\begin{tabular}{@{}|c|l|l|l|l|l|l|@{}}
\hline
\multicolumn{1}{|l|}{}                                                      & \multicolumn{1}{c|}{\begin{tabular}[c]{@{}c@{}}Problem\\ Domain(s)\end{tabular}}                                                                                      & \multicolumn{1}{c|}{\begin{tabular}[c]{@{}c@{}}Representation\\ Forms\end{tabular}}                                                                                                                                                                                                         & \multicolumn{1}{c|}{\begin{tabular}[c]{@{}c@{}}Completeness\\ Features\end{tabular}}                                   & \multicolumn{1}{c|}{\begin{tabular}[c]{@{}c@{}}Constraint\\ Transformation\end{tabular}}                                                                                                                                              & \multicolumn{1}{c|}{\begin{tabular}[c]{@{}c@{}}Related\\ Tasks\end{tabular}}                                                                                                                                    & \multicolumn{1}{c|}{\begin{tabular}[c]{@{}c@{}}Given\\ Examples\end{tabular}}                                                                                                                                                                                                                                               \\ \hline
\textbf{\begin{tabular}[c]{@{}c@{}}Conventional\\ Constraints\end{tabular}} & \begin{tabular}[c]{@{}l@{}}Within \\ optimization\\ domain\end{tabular}                                                                                               & \begin{tabular}[c]{@{}l@{}}Within \\ computational\\ representations with\\ structured and well\\ defined forms, {i.e.,}\\ equality and/or\\ inequality functions\end{tabular}                                                                                                                 & \begin{tabular}[c]{@{}l@{}}Fully known\\ constraint\\ representations\end{tabular}                                     & \begin{tabular}[c]{@{}l@{}}Transformation\\ only within the\\ computational\\ representations\end{tabular}                                                                                                                            & \begin{tabular}[c]{@{}l@{}}Transformation\\ into dual \\ problems\end{tabular}                                                                                                                                  & \begin{tabular}[c]{@{}l@{}}$g(\mathbf{x})=x_1^2+x_2=0$\\ $h(\mathbf{x})=x_1+x_2\geq 0$\end{tabular}                                                                                                                                                                                                                                                         \\ \hline
\textbf{\begin{tabular}[c]{@{}c@{}}Generalized\\ Constraints\end{tabular}}  & \begin{tabular}[c]{@{}l@{}}Covering \\a large \\ spectrum of \\ domains in AI\\modelings, \\ such as, 
\\mathematics,
\\cognition, \\ neuroscience, \\ psychology, \\linguistics,  \\social science, \\ physics, etc.\end{tabular} & \begin{tabular}[c]{@{}l@{}}Including natural\\ language \\ descriptions \\ and computational\\ representations \\ with (un)structured, \\ and/or ill (well) \\ defined forms, such \\ as, rule, graph, \\ table, functional, \\ equation, \\ (sub)model, \\ virtual data, etc.\end{tabular} & \begin{tabular}[c]{@{}l@{}}Including \\ fully and/or \\ partially \\ known\\ constraint\\ representations\end{tabular} & \begin{tabular}[c]{@{}l@{}}Possibly \\ requiring\\ transformations \\ between natural \\ language \\ descriptions \\ and \\ computational \\ representations \\ and/or from \\ unstructured \\ form into \\ structured one\end{tabular} & \begin{tabular}[c]{@{}l@{}}Possibly \\ requiring\\ constraint \\ mathematization, \\ constraint \\ coupling \\ selection,\\ parameter \\ identifiability, \\ and/or \\generalized\\ constraint \\ learning studies\end{tabular} & \begin{tabular}[c]{@{}l@{}}The system output \\ at the next step will \\ be a function of the  \\ current output, as \\ well as of the \\ output with a time \\ delay $\tau$ which is a \\ positive integer. \\ (Mathematization: \\ $x(t+1)=$ \\$f(x(t), x(t-\tau))$,\\ $\tau (\in {Z}^+)$ is unknown \\ for identification.)\\  \end{tabular} \\  \hline
\end{tabular}
\label{Table1}
\end{table*}

%
%
%
%

\subsection{Mathematical notation of generalized constraints }


In fact, the term GCs is not a new concept and it appeared in literature, such as a paper by Greene \cite{Greene:66:AIAA} in 1966. In the earlier studies, GC was used as a term without a formal definition until the studies by Zadeh \cite{Zadeh:86:AIMMS, Zadeh:96:TFS, Zadeh:11:FU} in 1986, 1996, and 2011, respectively. Zadeh presented a mathematical formulation of GCs in a canonical form \cite{Zadeh:96:TFS}: 
\begin{equation}\label{equ_2}
X~isr~R, 
\end{equation}
where $isr$ (pronounced “ezar”) is a variable copula which defines the way in which $R$ constrains a variable $X$. He stated that  ``{\it the role of R in relation to X is defined by the value of the discrete variable r}''. The values of $r$ for GCs are defined by probabilistic, probability, usuality, fuzzy set, rough set, etc; so that a wide variety of constraints can be included. Zadeh used GCs for proposing a new framework called ``{\it Generalized Theory of Uncertainty} (\textbf{GTU})''. He described that ``{\it The concept of a generalized constraint is the center piece of GTU}'' \cite{Zadeh:11:FU}. \textbf{The idea by Zadeh is very stimulating in the sense that we need to utilize all kinds of prior, or GCs, in AI modelings}. 

In 2009, Hu et al. \cite{Hu:09:IS} proposed so called ``{\it Generalized Constraint Neural Network} (\textbf{GCNN})'' model, in which the GCs considered by them were ``{\it partially known relationship} (\textbf{PKR})''  knowledge about the system being studied. Without awareness of the pioneer work by Zadeh, they proposed the formulation for a regression problem in a form of: 
\begin{equation}\label{equ_3}
\begin{array}{l}
min ~~~ ~~~~~~~~  \| y - g(\mathbf{x}, \mathbf{\theta} )\| \\
subject ~ to ~~~~g(\mathbf{x}, \mathbf{\theta} ) \propto R_i \langle f \rangle ; ~~ i = 1,2,...
\end{array}
\end{equation}
where $\mathbf{x}$ and $y$ are the input and output for a target function $f$ to be estimated, $g(\mathbf{x}, \mathbf{\theta} )$ is the approximate function with a parameter set $\mathbf{\theta}$, 
$R_i \langle f \rangle$ is the {\it i}th PKR of $f$, 
the symbol ``$\propto$” represents the term ``{\it is compatible with}'' so that ``{\it hard}'' or ``{\it soft}'' constraints can be specified to the PKR, and the symbol $\langle \cdot \rangle$ denotes ‘‘{\it about}'' because some PKRs cannot be expressed by mathematical functions. They adopted the term ``{\it relationships}'' rather than ``{\it functions}'', for the reason to express a variety of types of prior knowledge in a wider sense. Their work was inspired by the studies called ``{\it Hybrid Neural Network} (\textbf{HNN})'' models \cite{Psichogios:92:AIChE,Thompson:94:AIChE}, but adopted ``{\it generalized constraint}'' rather than ``{\it hybrid}'' as a descriptive term so that the mathematical meaning was clear and stressed.

\subsection{A view from mathematical spaces}
 
\begin{figure}
	\centering
	\includegraphics[width=8.0cm]{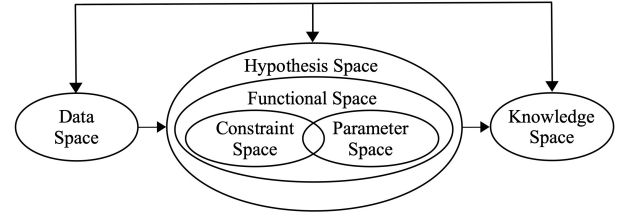}\\ 
	\caption{Schematic diagram of mathematical spaces studied in machine learning or artificial intelligence. The diagram is modified from \cite{Ran:14:NC} by adding the part of ``Constraint Space''.
	}
	\label{fig1}
\end{figure}

 Following the philosophy of Kalman \cite{Kalman:08:LEC}, we can view any study in machine learning (\textbf{ML}) or in AI is a mapping study among the mathematical spaces as shown in Fig. \ref{fig1}.

 The all spaces are interactive to each other, such as a {\it knowledge space} ({i.e.,} a set of knowledge) which is connected to other spaces via either an input  ({i.e.,} embedded knowledge) or an output ({i.e.,} derived knowledge) means. In fact, a knowledge space is not mutually exclusive with other spaces in the sense of sets. The diagram is given for a schematically understanding about the information flows in modelings as well as in an intelligent machine. The interactive and feedback relationships imply that intelligence is a dynamic process. 
 
 Most machine learning systems can be seen as a study on deriving a {\it hypothesis space} ({i.e.,} a set of hypotheses) from a {\it data space} ({i.e.,} a set of observations or even virtual datasets) \cite{Blockeel:01:EML}. The systems are also viewed as {\it parameter learning machines} if the concern is focused on a {\it parameter space} ({i.e.,} a set of parameters) \cite{Ran:14:NC}. This paper will focus on a {\it constraint space} ({i.e.,} a set of constraints) but highlight the problem of GCs in the context of mathematical modeling in AI. 
 
\subsection{ GCs as a new mathematical problem}
 
 AI machines generally involve optimizations \cite{Russell:10:Book}. However, most existing studies in AI concern more on CCs. There are primarily three types of constraints in CCs, that is,  equality, inequality, and integer constraints, respectively. All of them are given in a structured form. It is understandable that any modeling will involve the application of PI. In a real-world setting, PI does not always show it in a mathematical form of CCs. Therefore, in the designs of AI machines, we basically encounter a GC, rather than a CC, problem. 
 However, GCs are still considered as a new mathematical problem due to the following facts. 
 \begin{itemize}
 	\item From a mathematical viewpoint, we still miss ``{\it a mathematical theory of AI}'' if following the position of Shannon \cite{Shannon:48:BSTJ}. In other words, we have no theory to deal with GCs. For the given image, we can tell if a cat or a dog. This process is strongly related to the specific GCs embedded in our brains. However, for either deep learning or human brains, we are far away from understanding the specific GCs in a mathematical form. A theoretical study of GCs is needed, such as its notation and formalization as a general problem in AI. 
 	
 	\item From an application viewpoint, we need an applicable yet simple modeling tool that is able to incorporate any form of PI {\it maximally} and {\it explicitly} for understanding AI machines. Although application studies in AI encounter more GCs than CCs, not much studies are given under the notion of GCs. The GC problem is still far away from awareness for every related community, and mostly concerned within a case-by-case procedure. Moreover, the extra challenges listed in GCs have not been addressed systematically.
 \end{itemize} 
 


 
There are extensive literature implicitly covering GCs. The goal of this paper attempts to provide a {\it comprehensive} review about GCs so that readers will be able to understand the basics about them. 
The paper is not aiming at an extensive review of the existing works nor a rigorous about most terms and problems. For simplification of discussions, we will apply most terms directly and suggest to consider them in a broader sense. Only a few of terms are given or redefined for clarification. 
 We take a ‘‘{\it  top-down}'' way, that is, from outlooks to methods in the review. Therefore, in Section II, we will present outlooks given by different researchers, so that one can understand why GCs are critical in AI. Sections III and IV introduce the related methods in embedding GCs and extracting GCs, respectively. The summary and final remarks are given in Section V.

\section{AI machines and their futures}
\label{section2} 

This section will present overall understandings about AI machines and their futures. Because there exist numerous descriptions to the subjects, only a few of them are presented for the purpose of justifying the notion of GCs. 

\subsection{Five tribes and the Master Algorithm}
Domingos \cite{Domingos:15:Book} gave the systematic descriptions about the current learning machines and divided them with five ``{\it tribes}''. He suggest that the all tribes should be evolved into so called ``{\it the Master Algorithm}'' in future, which refers to a general-purpose learner.  
Table \ref{Table2} lists the five tribes with respect to the different features (or, GCs). Using a schematic plot, he illustrated the tribes by five circular sectors surrounding a core circle which is the Master Algorithm. He proposed a hypothesis below \cite{Domingos:15:Book}: 

``{\it All knowledge—past, present, and future—can be derived from data by a single, universal learning algorithm.}''    

The hypothesis suggests one important implication in future AI, that is, 
\begin{itemize}
	\item
     The future AI machines should be able to deal with any form of prior knowledge with a single algorithm. 
\end{itemize}

The implication presents the necessary conditions of future AI machines in regardless of the existence of the Master Algorithm. In fact, there exist other terms to describe future AI machines, such as ``artificial general intelligence ({\textbf{AGI})''\cite{Wang:07:AGI}. All of them have implied the utilization of GCs as a general tool.
\begin{table}
\caption{Five tribes within AI machines (from \cite{Domingos:15:Book})}
\begin{tabular}{|l|l|l|l|}
\hline
\textbf{Tribe} & \textbf{Representation}                                      & \textbf{Evaluation}                                              & \textbf{Optimization}                                               \\ \hline
Symbolists     & Logic                                                        & Accuracy                                                         & \begin{tabular}[c]{@{}l@{}}Inverse \\ Deduction\end{tabular}        \\ \hline
Connectionists & \begin{tabular}[c]{@{}l@{}}Neural \\ Networks\end{tabular}   & \begin{tabular}[c]{@{}l@{}}Squared \\ Errors\end{tabular}        & \begin{tabular}[c]{@{}l@{}}Gradient \\ Descent\end{tabular}         \\ \hline
Evolutionaries & \begin{tabular}[c]{@{}l@{}}Genetic \\ Programs\end{tabular}  & Fitness                                                          & \begin{tabular}[c]{@{}l@{}}Genetic \\ Search\end{tabular}           \\ \hline
Bayesians      & \begin{tabular}[c]{@{}l@{}}Graphical \\ Models\end{tabular}  & \begin{tabular}[c]{@{}l@{}}Posterior \\ Probability\end{tabular} & \begin{tabular}[c]{@{}l@{}}Probabilistic \\ Inference\end{tabular}  \\ \hline
Analogizers    & \begin{tabular}[c]{@{}l@{}}Supported \\ Vectors\end{tabular} & Margin                                                           & \begin{tabular}[c]{@{}l@{}}Constrained \\ Optimization\end{tabular} \\ \hline
\end{tabular}
\label{Table2}
\end{table}
 
\subsection{Reasoning methods}

In 1976, Box \cite{Box:76:JASA} illustrated the reasoning procedures (Fig. \ref{Box1}) for ``\textit{The advancement of learning}''. In Fig. \ref{Box1} (a), if the upper part and the lower part refer to data and knowledge, respectively, an \textit{induction} procedure follows a direction from data to knowledge and a \textit{deduction} procedure follows a direction from knowledge to data. He called them ``\textit{an iteration between theory and practice}''. In Fig. \ref{Box1} (b), he clearly showed that learning is an information processing involving both induction and deduction inferences as a dynamic system having a feedback loop. Usually, induction refers to a ``\textit{bottom-up}'' reasoning approach, and deduction a ``\textit{top-down}'' reasoning approach \cite{Trochim:01:Book}. Hence, the positions for upper part of the set ``\textit{Practice et al.}'' and the lower part of the set ``\textit{Hypothesis et al.}'' are better to be exchanged in Fig. \ref{Box1} (a), so that the meanings for the top and the bottom are correct. 

\begin{figure}
	\centering
\hspace*{-0.55cm}
	\includegraphics[scale=0.33]{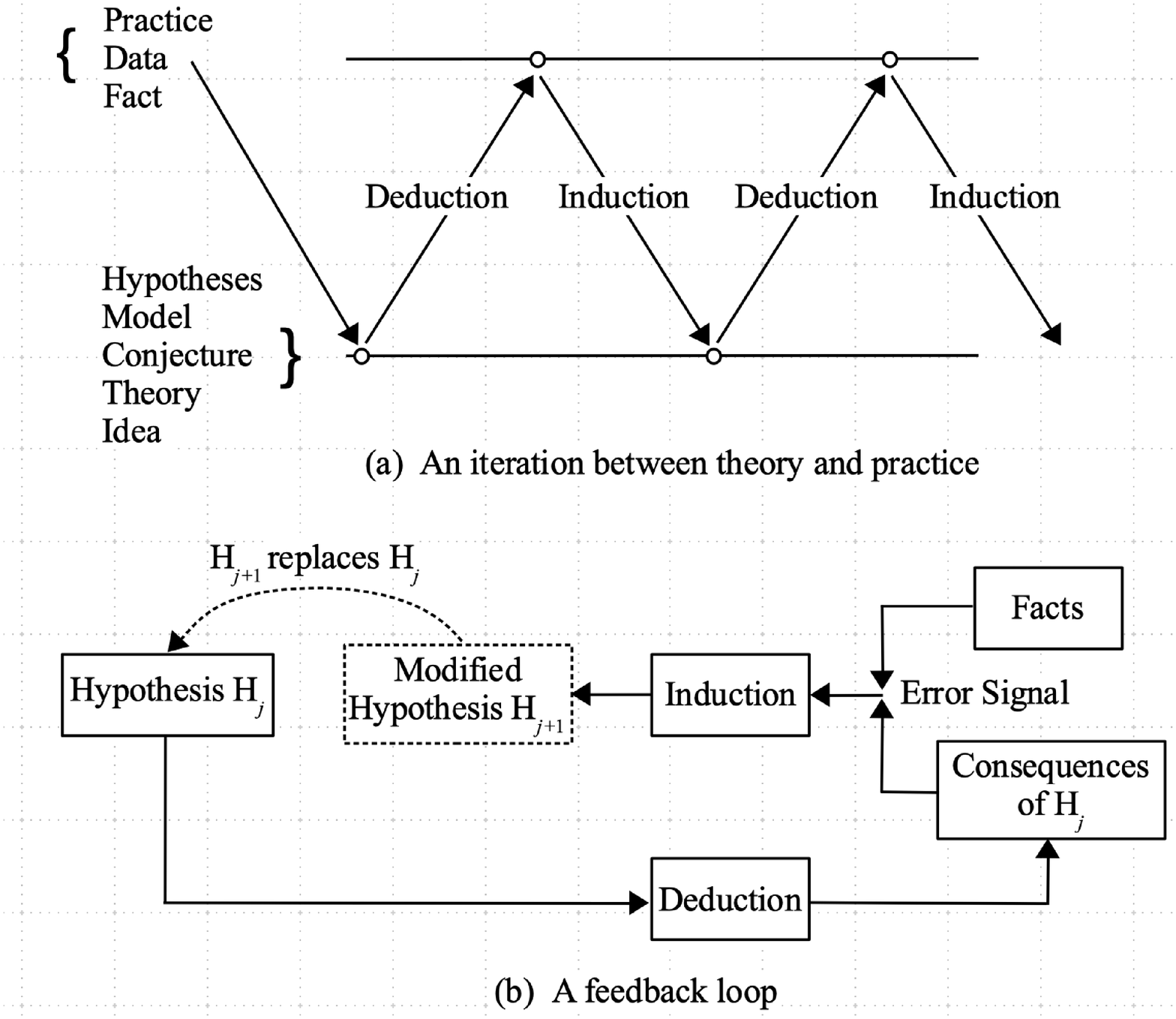}\\
	\caption{Schematic illustrations from \cite{Box:76:JASA} for ``\textit{The advancement of learning}''. 
	}
	\label{Box1}
\end{figure}

Hu et al. \cite{Hu:16:CCAA} adopted an image cognition example in \cite{Porter:54:AJP} for explanations of inferences. One can guess what it is about in {Fig. \ref{figPorter}}. If one cannot provide a guess, it is better to see {Fig. \ref{fig_A}} in Appendix, and then re-examine {Fig. \ref{figPorter}}. In general, most people can present a correct answer after seeing both figures. Only a small portion of people are able to tell the answer of {Fig. \ref{figPorter}} directly in the first-time seeing. 
\begin{figure}
	\centering
	\includegraphics[scale=0.18]{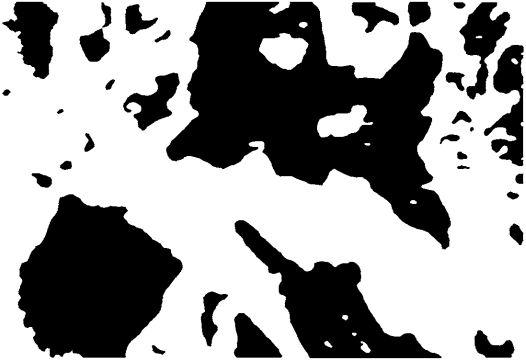}\\
	\caption{An image taken from \cite{Porter:54:AJP}. One may guess what it is about after carefully seeing this image.
	}
	\label{figPorter}
\end{figure}

If {Figs. \ref{figPorter}} and {\ref{fig_A}} represent the original data and prior knowledge respectively, the guessing answer of {Fig. \ref{figPorter}} is a {hypothesis}. We can understand a cognition process will involve both induction and deduction procedures. This is true for all people in either seeing or not seeing {Fig. \ref{fig_A}}. Generally, without human face prior in one's mind, the one is unable to provide a correct answer to the image in {Fig. \ref{figPorter}}. This cognition example confirms the proposal of Box \cite{Box:76:JASA} in Fig. \ref{Box1}. 
The study of Box and the cognition example provide the following implications in relation to GCs: 
\begin{itemize}
	\item
	The knowledge part in Fig. 2 (a) can be viewed as GCs, but how to formalize GCs in {Fig. \ref{fig_A}} is still a challenge.  
	\item The GCs are generally updated within a feedback loop, particularly when more data comes in. 
\end{itemize}

\subsection{Knowledge role in future AI machines}

Niyogi et al. \cite{Niyogi:98:IEEE} pointed out that ``{\it incorporation of prior knowledge might be the only way for learning machines to tractably generalize from finite data}". The statement suggested utilization of prior knowledge in a minimum sense. In a maximum sense, Ran and Hu \cite{Ran:14:NC} suggested the future AI machines should go to a higher position over ``\textit{Average Human Being}'' in both data utilization and knowledge utilization ({Fig. \ref{KD}}). When the term ``\textit{big knowledge}'' have been appeared, such as in \cite{Wu:16:AAS}, we provided a formal definition based on \cite{Hu:16:CCAA} below:

\begin{myDef} Big knowledge is a term for a knowledge base that is given on multiple discipline subbases, for multiple application domains, and with multiple forms of knowledge representations. 
\end{myDef}   

\begin{figure}
	\centering
	\includegraphics[scale=0.35]{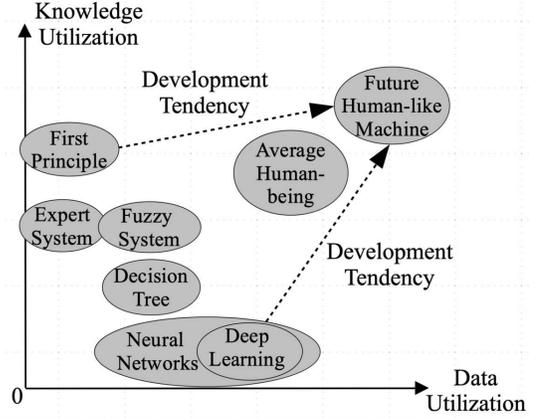}\\
	\caption{The relationship between current intelligent
		models and the future AI machines (modified from \cite{Ran:17:AAS}). The position for each model is given only in a non-rigorous sense. 
	}
	\label{KD}
\end{figure}

The knowledge role in future AI can be justified from the ultimate goals of AI. 
There are numerous understanding about the goals of AI in different communities. For example, Marr \cite{Marr:77:AI} described that ``\textit{the goal of AI is to identify and solve tractable information processing problems}''. Other researchers considered it as ``{\it passing the Turing Test}'' \cite{Saygin:00:MM}, General Problem Solver (\textbf{GPS}) \cite{Newell:76:CACM}, or AGI \cite{Wang:07:AGI,Goertzel:14:JAGI}. 
Some researchers proposed two ultimate goals, namely, scientific goal and engineering goal \cite{Brooks:94:AR, Wang:06:Book}. Following their positions and descriptions, we redefine the two goals of AI below: 
	\begin{itemize}
		\item Engineering goal: To create and use intelligent tools for helping humans maximumly for good.
		
		\item Scientific goal: To gain knowledge, better in depth, about humans themselves and other worlds.  
	\end{itemize}


When the scientific goal suggests an explicit role of knowledge as an output of AI machines, the engineering goal requires knowledge as an input in constructing the machines. For example, without knowledge, modelers will fail to reach the engineering goal in terms of ``{maximumly for good}''. 
From the discussions above, we can understand that GCs will play an important role in the studies of knowledge utilization, big knowledge, or realizing the ultimate goals of AI.

\subsection{Comprehension levels in AI machines}
Currently, AI machines are dominated by black-box DL models. For understanding AI machines as a rigorous science \cite{Doshi-Velez:17:arXiv}, more investigations have been reported, such as \cite{Ribeiro:16:SIGKDD, Lundberg:17:NIPS, Murdoch:19:arXiv, Alaa:19:NIPS, Yang:20:TNNLS}, in together with review papers, like \cite{Doran:17:arXiv, Guidotti:18:ACMCS, Mohseni:18:arXiv, Xie:20:arXiv}. In those investigations, several terms were given to define the different classes of 
AI machines, mostly based on application purposes. 
We present novel definitions below to the three specific classes of AI and show their relations in a mathematical notation.  
\begin{myDef} Transparent AI (\textbf{TAI}) refers to a class of methods in AI that are constructed by adding transparency over their counterparts without such operation. 
\end{myDef}   
\begin{myDef} Interpretable AI (\textbf{IAI}) refers to a class of methods in AI that are comprehensible for humans with interpretations from any natural language.
\end{myDef}
\begin{myDef} Explainable AI (\textbf{XAI}) refers to a class of methods in AI that are comprehensible for humans with mathematical explanations.	
\end{myDef}

\begin{figure}
	\centering
	\includegraphics[scale=0.18]{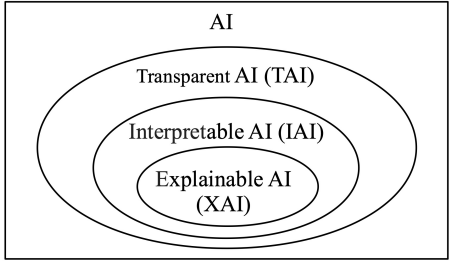}\\
	\caption{An Euler diagram of the three specific classes of AI machines.
	}
	\label{XAI2}
\end{figure}

Fig. \ref{XAI2} shows an Euler diagram of the different classes of AI machines. One can see their {\it strict subset} relations as   
\begin{equation}\label{equ_4}
\mbox{AI}  \supset \mbox{TAI} \supset \mbox{IAI} \supset \mbox{XAI}. 
\end{equation}
We can set the four classes of AI in an ordered way of \textit{comprehension (or knowledge) levels} from ``\textit{unknown}'', “\textit{shallow}”, “\textit{medium shallow}” to “\textit{deep}”, respectively. 
Their subset relations will remove the ambiguity in examination of methods, and provide a respective link to the comprehension levels in an ordered way.  A black-box AI  may be a preferred tool for some users. An autofocus camera for dummies is a good example. In this situation, AI tools with an unknown comprehension are fine for the users.

When TAI is a necessary condition of both IAI and XAI, we distinguish the both by their representation forms. A natural language representation is a naturally-evolved form for communication among humans. A mathematical representation is a structured form for communication among mathematicians. 
The differences of knowledge levels for IAI and XAI are due to the two forms of representations. We use Newton's second law as an example (Table \ref{EX}) to show the differences in representation forms and knowledge levels. The knowledge levels are given for a relative comparison. We can understand that a natural language representation will suffer the problems of {\it incompleteness}, \textit{ambiguity}, \textit{unclarity}, and/or \textit{inconsistency}. The problems will lead us for a shallow or incorrect understanding about the knowledge described. However, a mathematical representation will present a better form in describing knowledge exactly and completely. Although one is able to apply a natural language representation for describing Newton's second law exactly and completely, there exists no equality of sets for the two forms of representations. We need to note that GCs may be given in terms of either IXI or XAI, such as ``{\it different parts of perceptual input have common causes in the external world}'' and ``{\it mutual information}'', respectively, in processing adjacent patches of images \cite{Becker:92:NAT}.


\begin{table}
	\caption{An example of the relations between representation forms and knowledge levels.}
	\begin{center}
		\begin{tabular}{|c|c|c|c|}			
			\hline
			\begin{tabular}[x]{@{}c@{}}\textbf{Represen.}\\\textbf{Form}\end{tabular}& \begin{tabular}[x]{@{}c@{}}\textbf{Knowl.}\\\textbf{Level}\end{tabular} & \begin{tabular}[x]{@{}c@{}}\textbf{Unique. of }\\\textbf{Compre.}\end{tabular} & \textbf{Example}\\
			\hline
			\begin{tabular}[x]{@{}c@{}} {Natural}\\ {Language }\\ {Representation}\end{tabular} &\begin{tabular}[x]{@{}c@{}} {Medium }\\ {Shallow}\end{tabular}   &\begin{tabular}[x]{@{}c@{}} {No}\\ {or}\\ {Yes} \end{tabular} & \begin{tabular}[x]{@{}c@{}} {The acceleration}\\ {of an object is}\\ { proportional to the}\\ {  force imposed.}\end{tabular}  \\
			\hline	
			\begin{tabular}[x]{@{}c@{}} {Mathematical}\\ {Representation}\end{tabular}	 & Deep  &Yes & $\mathbf{F}=m\,\mathbf{a}$ \\
			\hline			
		\end{tabular}
		\label{EX}
	\end{center}
\end{table}

\section{Embedding GCs}

In a study of adding transparency to artificial neural networks ({\textbf{ANNs}), Hu et al. \cite{Hu:07:PRAI, Qu:11:TNN} described two main strategies, respectively:
	\begin{itemize}
		\item Strategy I: Embedding prior information into the models;
		\item Strategy II: Extracting knowledge or rules from the models;	
\end{itemize}
where a hierarchical diagram ({Fig. \ref{Qu11}}) was given so that one can get a simple and direct knowledge about the working format behind the method. The two strategies differ in the nature of the knowledge flow with respect to its model. Whereas the first strategy applies the available knowledge more like an input into the model, the second strategy obtains explicit knowledge as an outcome from the system. In this section, we will focus on the first strategy in the notion of GCs.  

\begin{figure*}
	\centering
	\includegraphics[width=.83\linewidth]{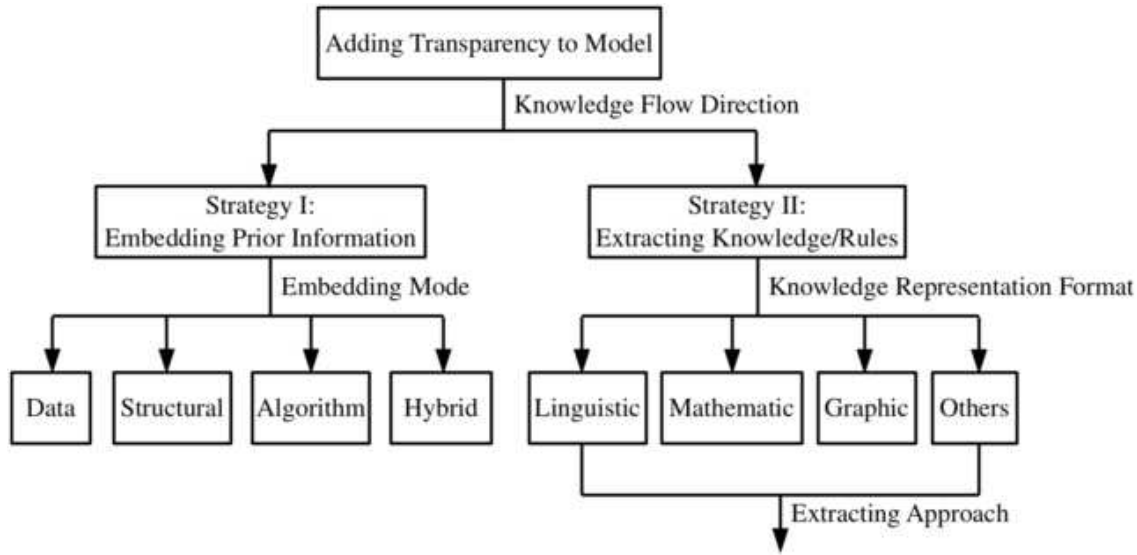}\\ 
	\caption{Hierarchical diagram of classifying methods in adding transparency to black-box models \cite{Hu:07:PRAI, Qu:11:TNN}.
	}
	\label{Qu11}
\end{figure*}

\subsection{Generic Priors in GCs}

Bengio et al. \cite{Bengio:13:PAMI} adopted a notion of ``{\it generic priors} (\textbf{GPs})'' for representation learning study in AI. They reviewed the works in lines of embedding the ten types of GPs into the models, that is: 
\begin{enumerate}
	\item smoothness, 
		\item multiple explanatory factors, 
		\item a hierarchical organization of explanatory factors, 
		\item semi-supervised learning, 
		\item shared factors across tasks, 
		\item manifolds, 
		\item natural clustering, 
		\item temporal and spatial coherence, 
		\item sparsity, 
		\item simplicity.	
\end{enumerate}

They also considered GPs to be ``{\it Generic AI-level Priors}''. We advocate this notion and attempt to provide its formal definition below: 
\begin{myDef} Generic priors (\textbf{GPs}) are a class of GCs which exhibits the generality in AI machines and is independent to the specific applications.  
\end{myDef}

Note that GPs may be given in a linguistic form so that they are not CCs. No fixed boundary exists between GPs and non-GPs (or called {\it specific priors}). Numerous studies have been reported on seeking GPs. Only some of them are given below with different backgrounds.

\subsubsection{Objective priors}
When using Bayesian tools, noninformative prior \cite{Jeffreys:46:RSL} and maximum entropy prior \cite{Bernardo:09:book} are suggested for realizing a higher degree of objectivity in reasoning.   

\subsubsection{Regularization}

When addressing an inverse problem, regularization is an important set of GPs to solve such {\it ill-posed} problem \cite{Tikhonov:63:RAS, Poggio:88:JC, Gori:18:Book}. When $L_1$ and $L_2$ penalties are used for the Lasso and ridge methods, they are corresponding to different GPs, {\it sparseness} \cite{Tibshirani:96:JRSS} and {\it smoothness} \cite{Bishop:93:TNN}, respectively. For tensor (or matrix) approximations, a regularization penalty can be a 
{\it low-rank}  \cite{Zhao:15:TNNLS}, or further GPs, such as, {\it structured  low-rank} with {\it nonnegative} elements \cite{Wang:18:TNNLS}. A semantic based regularization is expressed by a set of first-order logic (\textbf{FOL}) clauses \cite{Diligenti:12:ML}.

\subsubsection{Knowledge/fuzzy rules and GTU}
Knowledge or fuzzy rules \cite{Towell:94:AI,Ying:00:Book} provide a structured representation in embedding human semantic knowledge into the machines. Zadeh proposed a GTU framework \cite{Zadeh:96:TFS, Zadeh:11:FU} in order to enlarge GCs by covering other theoretical tools like rough sets \cite{Pawlak:91:Book}, belief functions \cite{Smets:93:IJAR}, etc.  

\subsubsection{Hints}
Abu-Mostafa \cite{Abu:93:NIPS} was one of pioneers of proposing GCs in AI, but used notions of ‘‘{\it intelligent hints}'' and ``{\it common sense hints}'' \cite{Abu:95:SA}. He summarized five types of hints, namely, 
\begin{enumerate}
	\item invariance, 
	\item monotonicity, 
	\item approximation, 
	\item binary, 
	\item examples, 
\end{enumerate}
in learning unknown functions, and developed a canonical representation of the hints \cite{Abu:93:NIPS}. He 
pointed out that ``{\it Providing the machine with hints can make learning faster and easier}'' \cite{Abu:95:SA}. The important implications of his study are: 
\begin{itemize}
	\item
	In AI modeling, one needs to apply various types of hints, or GCs, as much as possible ``{\it from simple observations to sophisticated knowledge}'' \cite{Abu:95:SA}.
	\item A mathematical and canonical representation of hints is necessary so that a learning algorithm is able to deal with hints numerically in a unified form.  
\end{itemize}

\subsubsection{GPs in unsupervised learning}

In \cite{Jain:10:PRL}, Jain presented several GPs in clustering studies, such as,
criteria based on the {\it Occam's razor principle} for determining the number of clusters (say, AIC, MDL, MML, etc), sample priors (say, {\it must-link}, {\it cannot-link}, {\it seeding}, etc). In unsupervised ranking of web pages in search engine studies, Google PageRank applied a semantic GP: ``{\it More important websites are likely to receive more links from other websites}'' \cite{Peretti:14:TR}, based on the link-structure data. When without such data in unsupervised ranking of multi-attribute objects, Li et al. \cite{Li:15:TKDE} suggested five GCs, in the notion of ``{\it meta rules}'', that should be satisfied for the ranking function, that is, 
\begin{enumerate}
	\item scale and translation invariance, 
	\item strict monotonicity, 
	\item compatibility of linearity and nonlinearity, 
	\item smoothness, 
	\item explicitness of parameter size. 
\end{enumerate}
The important implication of the studies in unsupervised ranking is gained below.
\begin{itemize}
	\item \textbf{Meta rules, or GPs, are important for assessing learning tools in terms of ``{\it high-level knowledge}'' \cite{Li:15:TKDE}, and become critically necessary when no ground truth (say, labels) exists}. 
\end{itemize}
\subsubsection{GPs in classification}
In \cite{Lauer:08:ML}, Lauer and Bloch presented a comprehensive review of incorporating priors into SVM machines. They considered two main groups of priors, namely, {\it class invariance} group and {\it knowledge on data} group. In \cite{Krupka:07:PMLR}, Krupka and Tishby took the notion of {\it meta features}, and listed several of them in association with the specific tasks. For example, the suggested GPs in the handwritten recognition are 
\begin{enumerate}
	\item position, 
	\item orientation, 
	\item geometric shape descriptors; 
\end{enumerate}
and in text classification are
\begin{enumerate}
	\item stem,  
	\item semantic meaning,
	\item synonym,  
	\item co-occurrence,
	\item part of speech, 
    \item is stop word. 
\end{enumerate}

\subsubsection{PKR in regression}

Considering a regression problem, Hu et al. \cite{Hu:09:IS} considered GCs in the notion of PKR. They listed ten types of PKR about nonlinear functions in a regression problem, that is,
\begin{enumerate}
	\item constants or coefficients,
	\item components,
	\item superposition,
	\item multiplication,
	\item derivatives and integrals,
	\item nonlinear properties,
	\item functional form,
	\item boundary conditions,
	\item equality conditions,
	\item inequality conditions.	  
\end{enumerate}
For example, the first type of PKR, a {\it physical constant}, was studied from the Mackey-Glass dynamic problem in a difference form:
\begin{equation}\label{IS_1}
\begin{array}{l}
x(t+1)= \frac {ax(t-\tau)} {1+x^{10} (t-\tau) } + (1-b) x(t).
\end{array}
\end{equation}
Within the observation data of the dynamics, one is able to have the prior for a {\it time delay} $\tau$, to be a positive, yet unknown, integer in the problem. They showed solutions to gain the estimations of physical constants $\tau$ and $b$ using GCNN model based on radius basis function (\textbf{RBF}) networks. In \cite{Qu:11:TNN}, Qu and Hu further studied ``{\it linear priors} (\textbf{LPs})'' within GCs, which was defined as ``{\it a class of prior information that exhibits a linear relation to the attributes of interests, such as variables, free parameters, or their functions of the models}''. The total 25 types of LPs were listed in regressions.        
	
\subsubsection{Virtual samples}
Virtual samples are one of the important GPs given in either an explicit or implicit form in modeling. A few of approaches are listed below.   
\begin{enumerate}
	\item {Virtual views} \cite{Niyogi:98:IEEE}: generating virtual views from a real view image,
	\item {Adding noise} \cite{An:96:NC, Lauer:08:ML}: adding noise types and levels into the input, output, and connection weights of ANNs for robust predictions, 
	\item {Graphical model} \cite{Jordan:04:SS, Koller:09:Book}: a framework of generating virtual data with probability, 
	\item SMOTE \cite{Chawla:02:AIR}: an approach of generating samples for the minority class in imbalanced classification, 
	\item {Adversarial examples} \cite{Goodfellow:14:NIPS, Yuan:19:TNNLS}: generating adversarial examples for better representation learning in deep neural networks.
\end{enumerate}


%
%
%

\subsection{Embedding modes and coupling forms}

When GCs are given, there exist numerous methods of 
incorporating them. Some researchers \cite{Hu:07:PRAI, Lauer:08:ML} suggested to catalog them with a few set of simple groups.
We take a notion of  ``{\it embedding modes}''  \cite{Qu:11:TNN} and definite it below.
 \begin{myDef} Embedding modes refer to a few and specific means of embedding GCs into a model. 
\end{myDef}

Hu et al. \cite{Hu:07:PRAI, Qu:11:TNN} listed the three basic embedding modes ({Fig. \ref{Qu11}}), namely, {\it data}, {\it algorithm}, and  {\it structural} modes, respectively. Their classification of the basic embedding modes is roughly correct since some methods may share the different modes simultaneously. 

Significant benefits will be gained from 
using the mode notion. First, one is able to reach a better understanding about numerous methods in terms of the simple modes. Second, each basic mode presents a different level of {\it explicitness} for knowing the priors embedded. Second, the best level is suggested to be the structural mode, then followed sequentially by the algorithm mode, and the data mode as the last \cite{Qu:11:TNN}. 

When Mitchell \cite{Mitchell:97:Book} described ``{\it three different methods for using prior knowledge}'', that is, ``{\it to initiate the hypothesis, to alter the search objective, and to augment search operators}'', we can say the three methods are basically within the {algorithm mode}.  Lauer and Bloch \cite{Lauer:08:ML} proposed a hierarchy of embedding methods with three groups, namely, ``{\it samples}'', ``{\it kernel}'',  and ``{\it optimization problem}''. The first group is the {data} mode and the last two are within the {algorithm} mode. For a better understanding, we list a few of approaches in terms of their embedding mode below. 

\subsubsection{Data mode} This is a basic mode of incorporating GCs into a model mostly from its data used. In apart from the methods in the virtual samples discussed previously, the other methods appeared, such as,  
\begin{enumerate}
	\item {Data with labels} \cite{Duda:01:Book} or {ranking list} \cite{Liu:11:Book},
	\item {Seeding data in learning} \cite{Basu:02:ICML},
	\item {Weight prior} \cite{Lecun:88:Proc},
	\item {Cost matrix in imbalance learning} \cite{He:09:TKDE},
	\item {Universum data} \cite{Weston:06:ICML},
	\item {Spatial context in image patches} \cite{Doersch:15:ICCV}.
\end{enumerate}
 
\subsubsection{Algorithm mode} This is a basic mode of incorporating GCs into a model mostly from its algorithm used. The methods mentioned previously, say, in objective prior or regularization, fall within this mode. The other methods can be: 
\begin{enumerate}
	\item {Objective functions and constraints} \cite{An:96:NC},
	\item {Activation functions or kernels} \cite{Haykin:99:Book, Shawe-Taylor:04:Book},
	\item {Weight initialization} \cite{Joerding:91:NN, Yam:00:NC}, 
	\item {Searching steps} \cite{Battiti:89:CS},
		\item {Meta learning} \cite{Vilalta:02:AIR},	
	\item {Transfer learning} \cite{Pan:09:TKDE},	
	\item {Curriculum learning and self-paced learning} \cite{Bengio:09:ICML, Kumar:10:NIPS},
	\item {Dropout} \cite{Hinton:12:arXiv}.
\end{enumerate}

\subsubsection{Structural mode} This is a basic mode of incorporating GCs into a model mostly from its structure used, such as
\begin{enumerate}
	\item {Decision tree} \cite{Quinlan:86:ML},
    \item {Neocognitron} \cite{Fukushima:80:BC}, 
	\item {Convolutional neural network} \cite{LeCun:89:NC},
	\item {Fuzzy system} \cite{Wang:96:Book}, 
	\item {Long short-term memory} \cite{Hochreiter:97:NC},	
	\item {Bayesian networks} \cite{Pearl:00:Book},
	\item {Markov networks} \cite{Li:09:Book},
	\item {Knowledge graph} \cite{Singha:12:TR}, 
\end{enumerate}

\subsubsection{Hybrid mode} This is a mixed mode by including at lest two basic modes for incorporating GCs into a model, such as 
\begin{enumerate}
	\item {First-principle + ANN models} \cite{Psichogios:92:AIChE},
	\item {Neuro-fuzzy models} \cite{Jang:96:Book},
	\item {Connectionist-symbolic models} \cite{Sun:95:Book,Townsend:20:TNNLS}, 
	\item {Graphical models} \cite{Jordan:04:SS},
	\item {SMOTE + C4.5} \cite{Chawla:02:AIR},
	\item {Markov logic networks} \cite{Richardson:06:ML}, 
	\item {GCNNs} \cite{Hu:09:IS}, 
	\item {Generative adversarial nets} \cite{Goodfellow:14:NIPS},
	\item {Graph Convolutional Network} \cite{Zhang:20:TKDE}.
	 
\end{enumerate}

Note that the classification above may be roughly correct to some approaches. For example, we put Bayesian networks 
within a structural mode to stress on its structural information, but put graphical models within a hybrid mode
to stress on its structural information and generation of virtual data. 
 
For coupling forms, we adopt the notion of ``Generalized Constraint (GC)'' model \cite{Ran:14:NC} for explanations. Fig. \ref{KDDM} 
depicts a GC model, which basically consists of two modules, namely, {\it knowledge-driven} (\textbf{KD}) submodel and {\it  data-driven} (\textbf{DD}) submodel. For simplifying the discussion, a time-invariant model is considered. A two-way coupling connection is applied between two submodels. For stressing on the modeling paradigm, a GC model is considered within the KDDM approach \cite{Fan:15:EM}. The general description of a GC model is given in a form of \cite{Ran:14:NC}:

 \begin{equation}\label{equ_5}
\begin{array}{l}
\mathbf{y} = \mathbf{f}(\mathbf{x}, \mathbf{\theta})= \mathbf{f_k}(\mathbf{x}, \mathbf{\theta_k}) \oplus \mathbf{f_d}(\mathbf{x}, \mathbf{\theta_d}),\\
\mathbf{\theta}=(\mathbf{\theta_k}, \mathbf{\theta_d}), ~ \mathbf{\theta_k} \cap \mathbf{\theta_d} = \emptyset,
\end{array}
\end{equation}
where $\mathbf{x}\in \mathcal{R}^n$ and $\mathbf{y}\in \mathcal{R}^m$ are the input and output vectors, $\mathbf{f}$ is a function for a complete model relation between $\mathbf{x}$ 
and $\mathbf{y}$, 
$\mathbf{f_k}$ and $\mathbf{f_d}$ are the functions associated to the KD and DD submodels, respectively. $\mathbf{\theta}\in \mathcal{R}^{(p+q)}$ is the parameter vector of the function $\mathbf{f}$, $\mathbf{\theta_k}\in \mathcal{R}^{p}$ and $\mathbf{\theta_d}\in \mathcal{R}^{q}$ are the parameter vectors associated to the functions $\mathbf{f_k}$ and $\mathbf{f_d}$, respectively. The symbol ``$\oplus$'' represents a coupling operation between the two submodels. 

\begin{figure}
	\centering
	\includegraphics[width=.61\linewidth]{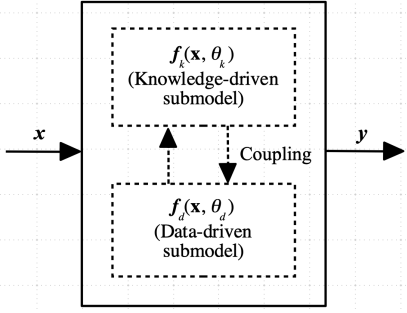}\\
	\caption{Schematic diagram of GC model including KD submodel and DD submodel \cite{Ran:14:NC, Fan:15:EM}. Two sets of parameters, $\theta_k$ and $\theta_d$ are associated with the two submodels, respectively. .
	}
	\label{KDDM}
\end{figure}

 \begin{myDef} Coupling forms refer to a few and specific means of integrating the KD and DD submodels when they are available. 
\end{myDef}

In fact, we can view the given GCs to be a KD submodel.
In 1992, Psichogios and Ungar \cite{Psichogios:92:AIChE} proposed an idea of using the first principle or empirical function to be KD submodels. When simulating a bioreaction process, they solved partial differential equations (\textbf{PDEs}) by a conventional approach except that a vector of physical parameters was estimated by a neural network submodel. Their HNN model applied the coupling between the two submodels in a {\it composition} form:
\begin{equation}\label{equ_P}
\begin{array}{l}
Composition ~ I:
{f}(\mathbf{x}, \mathbf{\theta})= {f_k}(\mathbf{x}, \mathbf{\theta_k}=\mathbf{f_d}(\mathbf{x}, \mathbf{\theta_{d}})),
\end{array}
\end{equation} 
where the prior was applied implicitly in which the physical parameter vector $\mathbf{\theta}$ was a nonlinear function to the input variables, rather than constant. We can call this form to be {\it Composition I (with parameter for learning)}, which can include a whole set or a part set of parameters for learning. This study is very enlightening to show an important direction to advance the modeling approach by the following implementations:
 \begin{itemize}
	\item Any theoretical or empirical model can be used as KD submodel so that the whole model keeps a physical explanation to the system or process investigated.  	
	\item The DD submodel, using either ANNs or other nonlinear tools, is integrated so that some unknown relations, nonlinearity, or parameters of the system investigated can be learnt.   
\end{itemize}

From the implementations above we can see that coupling form seems to be an extra challenge in the design of HNN or GC models. 
In 1994, Thompson and Kramer \cite{Thompson:94:AIChE} presented an extensive discussions about synthesizing HNN models for various types of prior knowledge. After combining a {\it parametric} KD submodel and a {\it nonparametric} DD submodel, they called the whole mode to be a {\it semiparametric} model. They considered two structures, {\it parallel} and {\it serial}, in coupling two submodels. Based on the two structures, Hu et al. \cite{Hu:09:IS} showed the four coupling forms in Fig. \ref{KDDM2}, and their mathematical expressions:
\begin{equation}\label{equ_Hu}
\begin{array}{l}
Superposition:~~~ 
\mathbf{f}(\mathbf{x}, \mathbf{\theta})= \mathbf{f_k}(\mathbf{x}, \mathbf{\theta_k}) + \mathbf{f_d}(\mathbf{x}, \mathbf{\theta_d}),\\
Multiplication:~~ 
\mathbf{f}(\mathbf{x}, \mathbf{\theta})= \mathbf{f_k}(\mathbf{x}, \mathbf{\theta_k}) \ast \mathbf{f_d}(\mathbf{x}, \mathbf{\theta_d}), \\
Composition ~ II:~~ 
\mathbf{f}(\mathbf{x}, \mathbf{\theta})= \mathbf{f_d}(\mathbf{f_k}(\mathbf{x}, \mathbf{\theta_k}), \mathbf{\theta_d}),\\
Composition ~ III:\,
\mathbf{f}(\mathbf{x}, \mathbf{\theta})= \mathbf{f_k}(\mathbf{f_d}(\mathbf{x}, \mathbf{\theta_d}), \mathbf{\theta_k}).
\end{array}
\end{equation} 

The four forms are simple and common in applications. In a parallel structure with a superposition form (Fig. \ref{KDDM2}(a)), when a KD submodel serves as a {\it core element} to simulate the main trends about the process investigated, a DD submodel works as an {\it error compensator} for uncertainties from the residual between the KD submodel output and the target output. One good example is a design of a nonlinear Kalman filter in \cite{Wilson:97:CCE}, where they applied a linear Kalman filter as a KD submodel and a neural network as a DD submodel to compensate for nonlinear deviations of the state variables. A multiplication form (Fig. \ref{KDDM2}(b)) was shown on a ``{\it Sinc}'' function in \cite{Hu:09:IS}:
\begin{equation}\label{IS_2}
\begin{array}{l}
f(x, y)= \frac {\sin {\sqrt {x^2+y^2}} } {x^2+y^2}, 
\end{array}
\end{equation} 
where the upper part was known, and the lower part was approximated by a RBF submodel. Hu et al. \cite{Hu:09:IS} used this example to show a possible problem introduced by a coupling which is discussed in the next section.

In a serial structure, another two composition forms are shown in eq. (\ref{equ_Hu}). The form of Composition II (Fig. \ref{KDDM2}(c)) corresponds to the inner function to be known as a KD submodel, such as a wavelet {\it preprocessor} before a neural network submodel in EEG analysis \cite{Kalayci:95:EMBM}.  
The idea of Composition III appeared in \cite{Jordan:92:CS} to force the output to be consistent with a {\it distal teacher}, like a KD submodel $\mathbf{f_k}(\mathbf{z}, \mathbf{\theta_k})$ in Fig. \ref{KDDM2}(d). Coupling forms can go quite complicated if considering more other structures, such as a {\it cyclic loop} in a dynamic process \cite{Chaffart:18:CCE}, or using a combination of various forms \cite{Stosch:14:CCE,Zendehboudi:18:AE}.    

\begin{figure}
	\centering
	\includegraphics[width=1.00\linewidth]{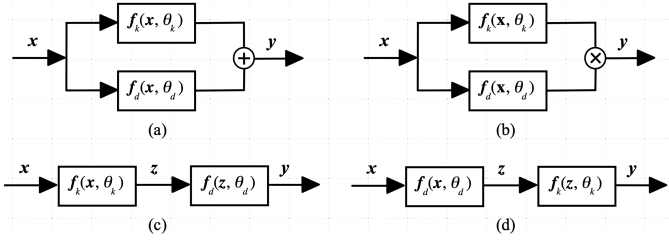}\\
	\caption{Four coupling forms of GC models \cite{Hu:09:IS}. Parallel structure: (a) Superposition, (b) Multiplication. Serial  structure: (c) Composition II (with inner function known), (d) Composition III (with outer function known).
	}
	\label{KDDM2}
\end{figure}

\subsection{Extra challenges in embedding GCs}

This subsection will discuss three extra procedures, or challenges, which are not discussed in the conventional ML study. More challenges exists, such as the mathematical conditions of {\it guaranteed better performance} for GC models over the models without using GCs \cite{Hu:09:IS}.  

\subsubsection{Constraint mathematization} GC can be given in any representation forms shown in Table \ref{Table1}. Actually, GCs in modeling are initially described by natural language. Therefore, in general, a procedure will be involved as defined below: 
\begin{myDef} Constraint mathematization refers to a procedure in modeling to transform GCs from natural language descriptions into mathematical representations. 
\end{myDef}

If GCs are given in a form of mathematical representations, the procedure above will be passed in modeling. However, when GCs are known only in a form of natural language descriptions, this procedure must be processed, like the given example in Table \ref{Table1}. The most difficulty in constraint mathematization is due to the problem called {\it semantic gap}. When various definitions about it exist from different contexts, such as from retrieval applications in \cite{Smeulders:00:PAMI}, we adopt the general definition:
\begin{myDef}[\cite{Hein:10:FPET}] ``{\it A semantic gap is the difference in meanings between constructs formed within different
representation systems.}''
\end{myDef}

In \cite{Hu:15:CIAC}, Hu suggested consider the semantic gap further by distinguishing two ways of transformations in modeling. A direct way is to transform natural language descriptions into mathematical representations. An inverse way is opposite to the direct one. 
Hence, the problem of semantic gap can be further extended by the two mathematical problems, namely, {\it ill-defined} and {\it ill-posed} problems. In \cite{Lynch:09:IJAIE}, Lynch et al. discussed various definitions of an ill-defined problem from the different researchers and proposed their definition below: 
\begin{myDef}[\cite{Lynch:09:IJAIE}] ``{\it A problem is ill-defined when essential concepts, relations, or solution criteria are un- or under-specified, open-textured, or intractable, requiring a solver to frame or recharacterize it. This recharacterization, and the resulting solution, are subject to debate.}''
\end{myDef}

Hadamard \cite{Hadamard:23:Book} was the first to define a mathematical term {\it well-posed} problems in mathematical modeling. Three criteria are given about {\it well-posed} definition to reflect three mathematical properties, namely, {\it existence}, {\it uniqueness} and {\it continuity}, respectively. Based on that, Poggio et al. proposed a simply definition of {\it ill-posed} problems below:
\begin{myDef}[\cite{Poggio:88:JC}] 
``{\it Mathematically ill-posed problems are problems where
the solution either does not exist or is not unique or does not depend
continuously on the data.}'' 
\end{myDef}

Any violation of one of the three properties will result in an ill-posed problem. Constraint mathematization works like a direct way of semantic gap, in which GCs may be ill defined. However, object recognition or constraint learning is an inverse way in which one generally encounters the problems with both ill-defined and ill-posed characterizations.     

In \cite{Diligenti:12:ML, Hu:16:ACL}, good examples were shown in constraint mathematization of first-order logic GCs for kernel machines and DNNs, respectively. However, it may be not easy to conduct constraint mathematization directly. For example, affective computing \cite{Picard:00:Book} will face more serious problems of semantic gap and ill-defined problems. The main difficulty sources mostly come from {\it ambiguity} and {\it subjectivity} of the linguistic representations about {\it emotional} or {\it mental} entities in modeling. In a study of emotion/mood recognition of music \cite{Rho:13:SC}, the linguistic terms were used, such as ``{\it valence}'' and ``{\it arousal}'' in Thayer’s two-dimensional emotion model \cite{Thayer:89:Book}, and emotion states like ``{\it happy}'', ``{\it sad}'', ``{\it calm}'', etc.  Many low-level and high-level features were used for establishing the relationship between music and emotion states. This study shows that we may need a model for constraint mathematization.



\subsubsection{Coupling form selection}
If a GC model (Fig. \ref{KDDM}) is used, a new procedure will appear in modeling as: 
\begin{myDef}
	Coupling form selection  (\textbf{CFS}) refers to a procedure in modeling to select a coupling form between the KD and DD submodels when they are available.
\end{myDef}

The subject of coupling form selection has not received much attention in ML or AI studies. Sometimes, the selection is determined by the problem given, like the multiplication form in approximation of ``{\it Sinc}'' function in eq. (\ref{IS_2}), where its upper part is known.  
In a general case, for the same GC or KD submodel, there may exist several, yet different, coupling forms to combine with a DD submodel. 
One example is about monotonicity of the prediction function with respect to some of the inputs. In \cite{Gupta:16:JMLR}, Gupta et al. presented 20 methods, including their own method. They divided those methods within four categories: A. constrain monotonic functions, B. post-process after training, C. penalize monotonicity violations when training, D. re-label samples to be monotonic before training. The four categories are better to be viewed in a structural mode with different coupling forms, such as Categories A and C are a superposition in Fig. \ref{KDDM2}(a), B in Fig. \ref{KDDM2}(d), and D in Fig. \ref{KDDM2}(c), respectively.  

\begin{figure}
	\centering
	\includegraphics[width=.93\linewidth]{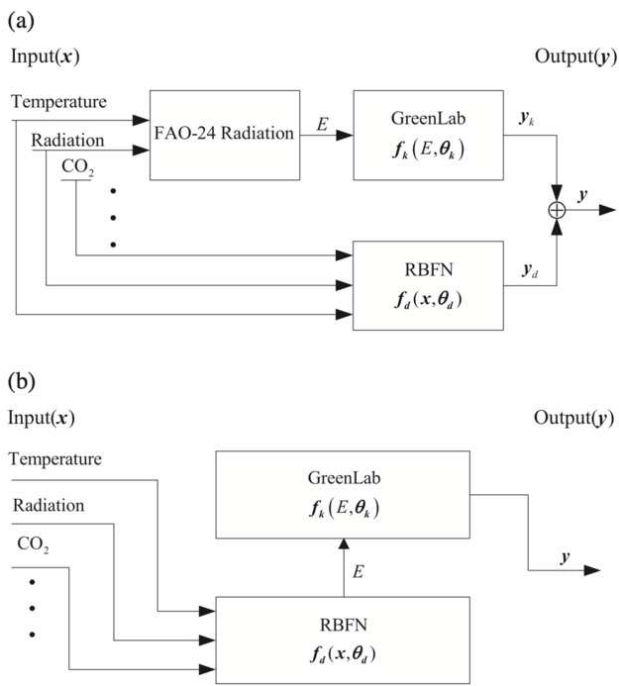}\\
	\caption{The knowledge-and-data-driven model (KDDM) with two cases of coupling: (a) superposition coupling operator and (b) composition coupling operator \cite{Fan:15:EM}.
	}
	\label{Fan15}
\end{figure}

In \cite{Fan:15:EM}, Fan {et al.} showed a study of CFS in their KDDM approach (Fig. \ref{Fan15}). When a KD submodel (called GreenLab) was given, they used RBF networks as a DD submodel. Two models, KDDM$_{sup}$ in Fig. \ref{Fan15}(a) and KDDM$_{com}$ in Fig. \ref{Fan15}(b), were investigated. The input variables were five environmental factors and the output was the total growth yield of tomato crop. 
After using the real-world data of tomato growth datasets from twelve greenhouse experiments over five years in a 12-fold cross-validation testing, KDDM$_{com}$ was selected because it showed a better prediction performance over KDDM$_{sup}$. In fact, KDDM$_{com}$ demonstrated a better interpretation about one variable, $E$, in plant growth dynamics than that of KDDM$_{sup}$. Their study also demonstrated several advantages of KDDM approach over the conventional KD or DD models, such as predictions of yields from different types of organs even some training data were missing.     

One can see that CFS is an extra challenge over the conventional modeling studies, but required to be explored systematically. We still need to define related criteria in the selection. When a performance is a main concern, some other issues may be also considered, such as learning speed, or interpretation of a model. \textbf{CFS presents a new subject in AI studies when more models or tribes are merged together as the Master Algorithm}. 

\subsubsection{Bio-inspired scheme for imposing constraints}
In the conventional ML study, the common scheme for imposing constraints is the Lagrange multiplier as a standard method \cite{Poggio:88:JC, Lecun:88:Proc, Gori:18:Book}. The question can be asked like  ``\textbf{{When ANNs emulate the synaptic plasticity functions of human brains, does a human brain apply the Lagrange multiplier if giving a new set of constraints}}''? Reviewing the image example of {Fig. \ref{figPorter}} again, if one relies on {Fig. \ref{fig_A}} for the correct recognition, we still do not know with which mathematical methods do our brains apply for describing and imposing the constraints.   

In \cite{Cao:16:IJCNN}, Cao {et al.} attempted to address this question based on the ``{\it Locality Principle (\textbf{LP})} '' \cite{Denning:72:ACM, Denning:05:ACM}. LP is originally come from classical physics in the {\it law of gravity} and states that an object is mostly influenced by its immediate surroundings. In computer science, Denning and Schwartz \cite{Denning:72:ACM} pointed out in 1972 that ``{\it Programs, to one degree or another, obey the principle of locality}'' and ``{\it The locality principle flows from human cognitive and coordinative behaviory}.'' In 2005, Denning further described that: ``{\it The locality principle found application well beyond virtual memory}'' and ``{\it Locality of reference is a fundamental principle of computing with many applications.}'' In neuroscience, LP can be supported from the knowledge: certain locations of a brain are responsible for certain functions (e.g. language, perception, etc.) \cite{Bear:10:Book}. Hence, Cao {et al.} \cite{Cao:16:IJCNN} stated that ``{\it All constraints can be viewed as memory. The principle provides both {time efficiency and energy efficiency}, which implies that constraints are better to be imposed through a local means.}'' They proposed a non-Lagrange multiplier method for equality constraints, falling within a locally imposing scheme (\textbf{LIS}). The main idea behind the proposed method was to make local modifications to the solutions after constraints were added. The method transformed equality constraint problems into unconstrained ones and solved them by a linear approach, so that convexity of constraints was no more an issue. The numerical examples were given on solving PDEs with Dirichlet or Neumann boundary constraints. The proposed method was possible to achieve an exact satisfaction of the equality constraints, while the Lagrange multiplier method presented only approximations. They further compared the two methods numerically on a one dimensional ``{\it Sinc}'', $f(x)=sin(x)/x$, with the interpolation constraints as: $f(0)=1$ and $f(\pi /2)=2/\pi$. The comparison aimed at ``{\it how to discover Lagrange multiplier method to be globally imposing scheme (\textbf{GIS}) or LIS?}'' They found the answers to be interpretation form depended. Suppose an interpretation form is given in the following expression:  
\begin{equation}\label{Cao16}
\begin{array}{l}
f(x)= f_{wc}(x) + f_m(x), 
\end{array}
\end{equation} 
where $f_{wc}(x)$ is the RBF output without constraints and $f_m(x)$ is the modification output after constraints are added in. From {Fig. \ref{Cao}}, one can see that $f_m(x)$ shows a global modification, or GIS, when using the Lagrange multiplier, but a local and smooth modification, or LIS, when using the proposed method. This interpretation form provided a locality interpretation from a ``{\it signal modification}'' sense. They also investigated the other interpretation form, but failed to give the clear conclusion from a  ``{\it synaptic weight changes}'' sense.  

\begin{figure}
	\centering
	\includegraphics[width=1.0\linewidth]{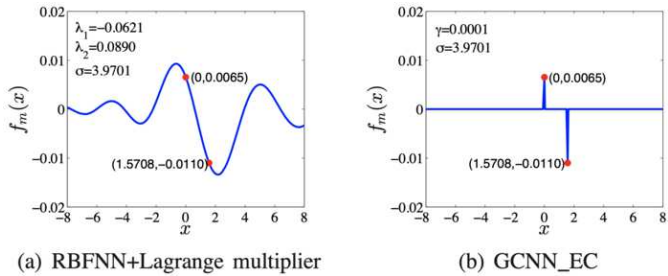}\\
	\caption{Modification plots, $f_m(x)$, for a {\it Sinc} function in which two constraints are added at $x = 0$ and $x = \pi/2$, respectively, (a) a global modification using the Lagrange multiplier and (b) a local and smooth modification using the proposed method \cite{Cao:16:IJCNN}.
	}
	\label{Cao}
\end{figure}

The study in \cite{Cao:16:IJCNN} is very preliminary, but shows an important direction for the future ML or AI studies by the following aspects:
\begin{itemize}
	\item
	\textbf{Locality principle is one of the important bases for realizing time efficiency and energy efficiency of bio-inspired machines}. We need to transfer principles, or high-level GCs, into mathematical methods in AI machine designs. Convolutional neural networks (\textbf{CNNs})  \cite{LeCun:89:NC} and RBFs \cite{Broomhead:88:CS} show the good examples in terms of the principle, satisfying a restricted region of the receptive field in biological processes \cite{Hubel:68:JP}. More investigations are needed to explore LP in a wider sense for designs, such as LIS on the inequality constraints, or its role in {\it continuous learning} with another important GC: {\it no catastrophic forgetting} (\textbf{NCF}) \cite{McCloskey:89:PLM,Kirkpatricka:17:NAS}. 
	\item \textbf{The hypothesis and selection of LIS and/or GIS suggest a new subject in the study of brain modeling or bio-inspired machines}. When LP is rooted on the classical physics, quantum mechanics might violate locality from the entanglement phenomenon \cite{Einstein:35:PR}. The different studies have been reported about brain modeling based on quantum theory, such as {\it quantum mind} \cite{Hameroff:94:JCS, Koch:06:NAT} and {\it quantum cognition} \cite{Busemeyer:12:Book}. Those studies suggested that consciousness should be a {\it quantum-type} prior, and should be treated with non-classical methods. The subject shows that LP or {\it nonlocality} will be a main issue for realizing a brain-inspired machine, and GCs should be treated according to their principle behind.  
\end{itemize}

\section{Extracting GCs}

In 1983, Scott \cite{Scott:83:AAAI} examined
the nature of learning and presented the definition below: 

``{\it Learning is any process through which a
	system acquires synthetic a posteriori
	knowledge.}''

The definition exactly reflects the purpose of learning in AI systems. \textbf{After {\it a posterioi} knowledge is acquired, AI systems are better to take it as {\it a priori} so that the systems are able to advance along with explicit knowledge updated and increased through an automatic or interactive means}. 

Extracting knowledge is a main concern in the areas of knowledge discovery in databases (\textbf{KDD}) and data mining \cite{Cios:98:Book, Bengio:00:TNN, Mitra:02:TNN, Han:11:Book}. It is also an important strategy in the context of adding transparency of ANN modeling \cite{Hu:09:IS} or IAI/XAI \cite{Murdoch:19:arXiv}. {Fig. \ref{Qu11}} shows a selection of representation formats (or forms) to be a first procedure within knowledge (or GCs) extracting approaches. Three basic forms are listed, namely, {\it linguistic}, {\it mathematic} and {\it graphic}, but the mathematic form is considered in a narrow sense here for not including the other two forms. The {\it other} form can be a non-formal language or a combination of the basic forms. This classification is roughly correct for the purpose of distinguishing the knowledge forms about extracting approaches. Furthermore, extracting GCs can be viewed as another subject defined below:
 \begin{myDef} Generalized constraint learning (\textbf{GCL}) refers to a problem in machine learning to obtain a set of generalized constraint(s) which is embedded in the dataset(s) and/or the system(s) investigated.
\end{myDef}

GCL is an extension of {\it prior learning}
(\textbf{PL}) \cite{Zhu:97:PAMI}, 
{\it constrained based approach}
(\textbf{CBA}) \cite{Gori:18:Book}, and
{\it constraint learning} (\textbf{CL}) \cite{DeRaedt:18:AAAI}. 
It stresses GCs from both data and system sources. \textbf{GCL suggests that AI machines should serve as a tool for scientific discovery \cite{Karpatne:17:TKDE}, such as on human brains}. Supposing that a recognition mechanism, or an objective function, is fixed in a brain, GCs will explain why we identify a man, rather than a woman, from {Fig. \ref{figPorter}}. In the followings, we review the extracting approaches, or GCL, according to the representation formats in Fig. \ref{Qu11}. 


\subsection{Extracting linguistic GCs}

The form of linguistic representation of knowledge considered here is symbolic rules, that is,  
\begin{equation}\label{if_rule}
\begin{array}{l}
Rule: ~ \mbox{If} ~<condition> ~\mbox{Then} ~<result> .
\end{array}
\end{equation} 

In 1988, Gallant \cite{Gallant:88:ACM} was a pioneer of extracting rules using ANN machines, for the purpose of generating an expert system automatically from data. He called such machine {\it connectionist expert system} (\textbf{CES}). The rules extracted were conventional (Boolean) symbolic ones. This work was very stimulating by showing a novel way for ANN modeling to obtain {\it explicit knowledge}:
\begin{equation}\label{}
\mbox{ Data~ $\rightarrow$ ~ANNs $\rightarrow$ ~CES $\rightarrow$}
\begin{array}{l}
\mbox{Explicit knowledge} \\ \mbox{with extracted rules} ,
\end{array}
\end{equation} 
in comparison with a conventional way:
\begin{equation}\label{equ_sinc}
\mbox{ Data~ $\rightarrow$ ~ ANNs ~$\rightarrow$}
\begin{array}{l}
\mbox{Implicit knowledge} \\ \mbox{with parameters}.
\end{array}
\end{equation} 
According to Haykin \cite{Haykin:99:Book}, ANNs gain knowledge from data learning and store the knowledge with its model parameters, i.e. {\it synaptic weights}. Dienes and Pemer \cite{Dienes:96:IC} considered that ANNs produce a type of {\it implicit knowledge} from those parameters.

After the study of Gallant \cite{Gallant:88:ACM}, a number of investigations \cite{Wang:92:SMC,Towell:93:ML, Fu:15:TSMC, Tsukimoto:00:TNN,Zhou:03:AIC,Kolman:05:TNN, Tran:16:TNNLS,Townsend:20:TNNLS} appeared on rule extraction from ANNs. 
In \cite{Wang:92:SMC}, Wang and Mendel developed an approach of generating fuzzy rules from numerical data. Fuzzy rules are a linguistic form to represent vague and imprecise information. In their fuzzy model, its fuzzy rule base was able to be formed by experts or rules extraction from the data. Fuzzy models are the same as ANNs in the sense of {\it universal approximator} \cite{Hornik:89:NN, Wang:96:Book}, but beneficial with regards to rules (or linguistic GCs) in modeling. In \cite{Tsukimoto:00:TNN}, Tsukimoto extracted rules applicable for both continuous and discrete values from either recurrent neural network (\textbf{RNN}) or multilayer perceptron (\textbf{MLP}). Several review papers \cite{Andrew:95:KS,Mitra:00:TNN,Jacobsson:05:NC} reported on rule extractions and presented more different approaches.

In evaluation of ANN based approaches, Andrew {et al.} \cite{Andrew:95:KS} provided a taxonomy of five  features. Among the features, the one called {\it quality} seems mostly important, for which they proposed four criteria in rule quality examinations, namely,  
{\it accuracy}, {\it fidelity}, {\it consistency}, and {\it comprehensibility}. They suggested the measurements of the criteria. This work shows an importance to evaluate quality of extracted GCs from different aspects. 

\subsection{Extracting mathematic GCs} 
In this subsection, we refer mathematic GCs in a narrow sense without including the forms of linguistic and graphic representations. We present several investigations which fall within the subjects of extracting mathematic GCs. 

\subsubsection{GCL for unknown parameters or properties}
In \cite{Hu:09:IS}, a GC was given first in a linguistic form, as shown by the GCs example in 
Table 1. This example describes the Mackey-Glass dynamic process, shown in eq. (\ref{IS_1}). The linguistic GC (or hypothesis) was finally transformed into a mathematic form as
\begin{equation}\label{equ_GC}
x(t+1)=f\{(x-\tau)\}+(1-b)x(t).
\end{equation} 
For the given time series dataset with $\tau = 17$ and $b = 0.1$, Hu et al. \cite{Hu:09:IS} used their GCNN model for the prediction and also obtained the solutions $\hat\tau  = 17$ and $\hat b  = 0.1096$. 

In the same paper, Hu et al. also used GCNN as a {\it hypothesis tool} to discover the hidden property of data in approximating ``{\it Sinc}'' function,  eq. (\ref{IS_2}). For the given GC, a GCNN model exhibited an abrupt change in the output $f$ when $x=y=0$. They suggested the hypothesis of ``{\it removable singularity}'' for the abrupt change, and confirmed it from the data and analysis procedures. Their study showed that ML models can be a tool for identifying unknown parameters and hidden properties of physical systems.   

\subsubsection{GCL for unknown functionals}
Recently, Alaa and van der Schaar \cite{Alaa:19:NIPS} proposed a novel and encouraging direction to extract mathematic GCs in form of analytic functionals from ANN models. In their study, a black-box ANN model was described by $f(\mathbf{x})$. They formed a symbolic metamodel $g(\mathbf{x})$ using Meijer $G$-functions. The class of Meijer $G$-functions includes a wide spectrum of common functions used in modeling, such as,
polynomial, exponential, logarithmic, trigonometric, and Bessel functions. Moreover, Meijer $G$-functions are analytic and closed-form functions, even for their differential forms. By minimizing a ``{\it metamodeling loss}'' $l(g(\mathbf{x}), f (\mathbf{x}))$, they obtained $g(\mathbf{x})$ with a parameterized representation of symbolic expressions. They conducted  numerical experiments on four different functions as the given GCs, that is, exponential, rational, Bessel and sinusoidal functions, respectively. Their approach was able to figure out the first three functional forms among the four functions. 

In Fig. 1, we show that functional space is among the primary concerns in ML or AI. When the true functional form is unknown to the system investigated, modelers will involve a subject defined below:
\begin{myDef}
	Functional form selection (\textbf{FFS}) refers to a procedure in modeling to select a suitable functional form for describing the system investigated from a set of candidate functionals.
\end{myDef}

FFS has received little attention in AI communities. For simplification, most of the existing AI models directly apply a preferred functional without involving FFS. If considering AI to be a discovering tool, FFS should be considered first in modeling, particularly to complex processes, say, in economics \cite{Griffin:87:WJAE}.
In view of system identification \cite{Nelles:13:Book, Schoukens:19:TMCS}, {FFS} will be more basic and difficult than {\it parameter estimations} in nonlinear modeling. The study in \cite{Alaa:19:NIPS} shows an important ``{\it gateway}'' in FFS, as well as in obtaining more fundamental knowledge, functional forms, about physical systems. Similar to the subject of CFS, FFS shows another open problem requiring a systematic study for ML or AI communities. 

\subsubsection{Parameter identifiability}
Bellman and Åström \cite{Bellman:70:MB} proposed the definition of {\it structural identifiability} (\textbf{SI}) in modeling. The term ``structural'' means the internal structure of a model, so that SI is independent of the data applied. Because any ML model with a finite set of parameters can be viewed as a ``{\it parameter learning machine}'' \cite{Ran:14:NC}, Ran and Hu adopted the notion of {\it parameter identifiability} to be the theoretical uniqueness of model parameters in statistical learning \cite{Ran:17:NC}. Yang {et al.} \cite{Yang:08:NC} showed an example why parameter (or structural) identifiability becomes a prerequisite before estimating a physical parameter in GCNN model (Fig. \ref{KDDM2}(b)) in a form of: 
\begin{equation}\label{Yang}
y= f_k(x,\mathbf{\theta_k}) \times f_d(x,\mathbf{\theta_d})= e^{-\alpha x} \times f_d(x,\mathbf{\theta_d}),
\end{equation} 
where damping coefficient $\alpha$ ($>0$) is a single physical parameter in the KD submodel $f_k(x,\mathbf{\theta_k})$. They demonstrated that $\alpha$ is unidentifiable if $f_d(x,\mathbf{\theta_d})$ is a RBF submodel, which suggests no guarantee of convergence to the true value of $\alpha$. However, $\alpha$ becomes identifiable if a sigmoidal feed-forward network (\textbf{FFNs}) submodel is applied. In a review paper \cite{Ran:17:NC}, Ran and Hu presented the reasons of identifiability study in ML models:

 ``{\it Identifiability analysis is important not only for models whose parameters have physically interpretable meaning, but also for models whose parameters have no physical implications, because identifiability has a significant influence on many aspects of a learning problem, such as estimation theory, hypothesis testing, model selection, learning algorithm, learning dynamic, and Bayesian inference.}'' 
 
 For example, Amari {et al.} \cite{Amari:06:NC} showed that FNNs or Gaussian mixture models (\textbf{GMMs}) will exhibit very slow dynamics of learning due to the nonidentifiability of parameters from plateaus of neuromanifold. Hence, \textbf{parameter identifiability provides mathematical understanding, or explainability, about learning machines in terms of their parameter spaces involved}. For details about parameter identifiability in ML, see \cite{Paulino:94:JISS, Koller:09:Book, Ran:17:NC} and the references therein.

\subsection{Extracting graphic GCs}
Human is more efficient in gaining knowledge from graphic means than from other means, such as from linguistic one. This paper considers graphic GCs in a wider sense. We refer graphic GCs to be GCs described by a graphic representation as well as shown by its graphic form. Hence, this subsection also includes graphic approaches for extracting GCs.  

\subsubsection{Data visualization for GCs}
Data visualization (\textbf{DV}) is a graphic representation of data for modelers/users to understand the features of the data. In ML studies, 
DV is one of the most common techniques in modeling so that modelers are able to gain the related GCs for the design guidelines as well as for the interpretation to the response of models \cite{Turkay:17:MLKE}.  
This technique can extend to the subjects of {\it information visualization} (\textbf{IV}) \cite{Liu:14:VC} or {\it visual analytics} (\textbf{VA}) \cite{Endert:17:CGF}. Some DV tools are able to present knowledge from data in a mathematic form, such as histogram, heatmap, etc. We present an example below to show that this is not a case in general. \textbf{DV tools usually exhibit knowledge in a form of GCs, which require users to seek and abstract}. Constraint mathematization will be another challenge in abstractions of knowledge. 

When high-dimensional data are difficult to interpret, researchers developed many approaches for viewing their intrinsic structures in a low-dimensional space \cite{Geng:05:SMCB}.
In \cite{Maaten:08:JMLR}, van der Maaten and Hinton proposed an approach called t-SNE (t-Distributed Stochastic Neighbour Embedding) for visualization of high-dimensional data in a 2-D or 3-D space. Different from a PCA approach, t-SNE is a nonlinear dimensionality reduction technique. For the handwritten digits 0 to 9  in the MNIST dataset, they showed the 2-D visualization results with t-SNE ({Fig. \ref{T-SNE}}). The raw data is given in high dimensions with a size of 784 ($28 \times 28$ pixels). The t-SNE plot of {Fig. \ref{T-SNE}} clearly demonstrates the hidden patterns of clusters in the high dimensional data. The patterns present a rather typical form of GCs instead of CCs. Modelers need to summarize them in form of either natural languages or mathematical descriptions,  such as {\it visual outliers} \cite{Rauber:17:TVCG} in classifications, or {\it semantic emergence}, {\it  shift}, or
{\it death} in dynamic {word embedding} \cite{Yao:18:ACM} from t-SNE plots. 

\begin{figure}
	\centering
	\includegraphics[width=.82\linewidth]{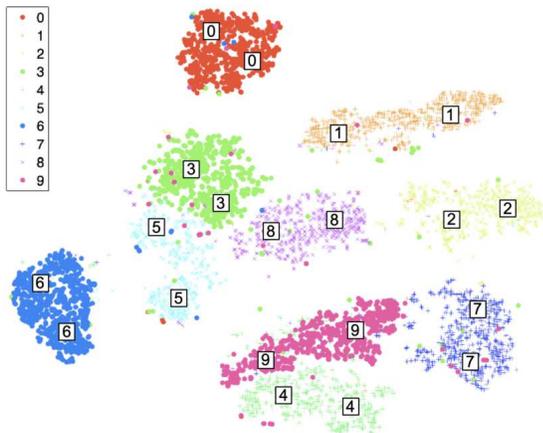}\\
	\caption{Visualization results with t-SNE on the MNIST dataset (Modified based on \cite{Maaten:08:JMLR} by adding labels to the clusters).
	}
	\label{T-SNE}
\end{figure}

The most challenge in visualization is to develop/seek a DV tool suitable for revealing intrinsic properties, or theoretical understanding, about data. One good example is a study by Shwartz-Ziv and Tishby \cite{Shwartz-Ziv:17:arXiv} in proposing a DV tool, called {\it information plane}. This tool presented a novel visualization of data from information-theoretic understanding, and was used for explaining the structural properties and leaning properties of DNNs.  

\subsubsection{Model visualization for GCs}
Model visualization (\textbf{MV}) is a graphic representation of ML/AI models about their architectures, learning processes, decision patterns, or related entities for modelers/users to understand the properties of the models. MV is another means to foster insights into models, and it may often involve techniques from DV, such as learning trajectories in optimizations \cite{Gallagher:03:SMCB}, or a heatmap for explaining convolutional neural networks (\textbf{CNNs}) \cite{Zhou:16:CVPR,Samek:16:TNNLS}.

If regardless of a common representation of architecture diagrams of models, 
it was reported that Hinton diagrams \cite{Hinton:86:PDP} was among the first tools in visualizing ANN models \cite{Craven:92:AIT}. This tool was designed for viewing connection strengths in ANNs, using color to denote sign and area to denote magnitude. With the help of Hinton diagram, Wu {et al.} \cite{Wu:96:CILS} gained the knowledge about unimportant weights for pruning ANN architectures. In 1992, Craven and Shavlik \cite{Craven:92:AIT} reviewed a number of visualization techniques for understanding decision-making processes of ANNs, such as Hinton diagrams \cite{Hinton:86:PDP}, bond diagram and Trajectory diagrams \cite{Wejchert:90:NIPS}, etc. They also developed a tool, called Lascaux, for visualizing ANNs. Graphic symbols were used to show the connections and neurons with activation and error signals, so that modelers were able to gain some insight into the learning behavior of ANNs. In the early study of MV, some researchers proposed a study of ``{\it Illuminating or opening the black box}'' of ANN models \cite{Olden:02:EM, Tzeng:05:TV}. In \cite{Olden:02:EM},  Olden and Jackson applied neural interpretation diagram (\textbf{NID}), Garson's  algorithm \cite{Garson:91:AIE}, and sensitivity analysis, so that variable selections can be understood by visual perception of modelers. In \cite{Tzeng:05:TV},  Tzeng and Ma used a new MV for knowing the working architectures of ANNs when the voxel of an image was corresponding to the brain material or not. The different architectures provided the relational knowledge between the important variables and voxel types. 

When deep learning (\textbf{DL}) models are emerged, more MV studies are reported on understanding the inner-workings of this type of black-box models \cite{Yosinski:15:ICML, Kahng:18:TVCG}. In \cite{Liu:17:TVCG}, Liu {et al.} developed a MV tool for visualizing the deep CNNs, called CNNVis. {Fig. \ref{Liu17}} shows a CNN model including four groups of convolutional layers and two fully connected layers. The model consisted of thousands of neurons and millions of connections in its architecture. Using CNNVis to simplify large graphs, they were able to show representations in clusters  for better and fast understanding about pattern knowledge of CNN models. They conducted case studies on the influence of network architectures and diagnosis of a failed training process to demonstrate the benefits of using CNNVis. In a different GC form, Zhang {et al.} \cite{Zhang:19:CVPR} applied decision trees in a semantic form to interpret CNNs models. For the other DL models, one can refer to the studies on recurrent neural networks (\textbf{RNNs}) \cite{Karpathy:16:ICLR, Strobelt:17:TVCG} and reinforcement learning (\textbf{RL}) \cite{Greydanus:18:ICML,Rupprecht:20:ICLR}.
Some review papers \cite{Zhang:18:FITEE, Kahng:18:TVCG, Hohman:19:TVCG} appeared recently about visualization of DL models. In general, MV techniques provide intuitive GCs in an unstructured form, and present a human-in-the-loop interactive of AI \cite{Hohman:19:TVCG}, such as to modify models or to adjust learning processes \cite{Endert:17:CGF, Strobelt:18:TVCG}. 

\begin{figure*}
	\centering
	\includegraphics[width=0.88\linewidth]{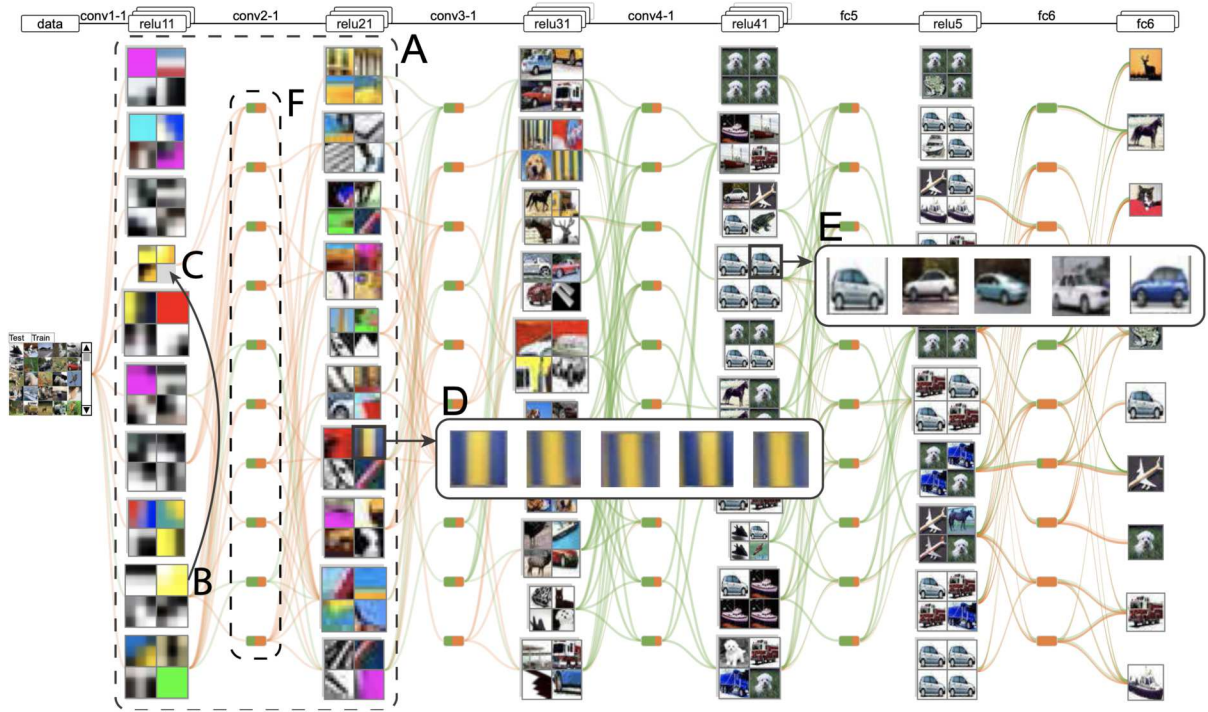}\\ 
	\caption{Diagram of CNNVis for visualizing deep convolutional neural networks \cite{Liu:17:TVCG}, in which one is able to see the pattern flow in clusters.
	}
	\label{Liu17}
\end{figure*}

\begin{figure*}
	\centering
	\includegraphics[width=1.0\linewidth]{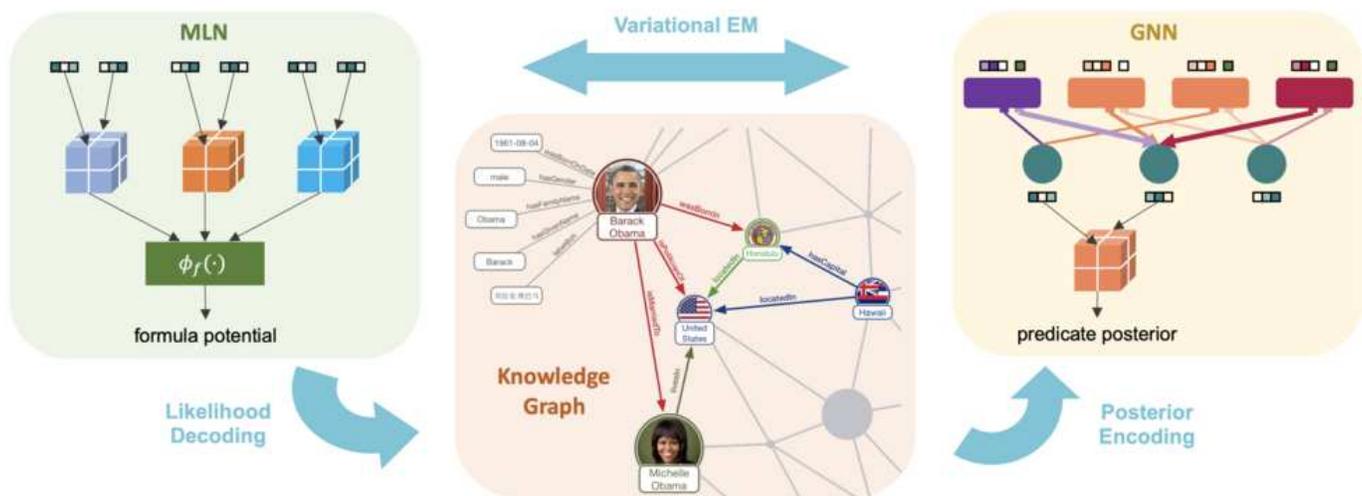}\\ 
	\caption{Diagram of ExpressGNN for combining MLN and GNN using the variational EM framework \cite{Zhang:20:ICLR}.
	}
	\label{Zhang20}
\end{figure*}

\subsubsection{Discovering GCs from graphical models} In general, we can say that \textbf{graphical models (\textbf{GMs}), using graphic representations, are the best and direct means to describe worlds, either physical or cyber one}. The main reason is from the facts below: (i) most objects are distinguished from their geometries, and (ii) they may be linked internally by their own entities or externally to each others. For every objects or data, their underlying graph structures are the most important knowledge we need to explore \cite{Scarselli:09:TNN}. Buntine \cite{Buntine:91:UAI} pointed out that ``{\it Probabilistic graphical models are a unified qualitative and quantitative framework for representing and reasoning with probabilities and independencies}'', and ``{\it  are an attractive modeling tool for knowledge discovery}''. We will present two sets of studies below to show that the knowledge discovered in GMs is often given in a form of GCs, instead of CCs.  

In \cite{Zhang:20:ICLR}, Zhang {et al.} developed a tool, called ExpressGNN, by combing Markov logic networks (\textbf{MLNs}) \cite{Richardson:06:ML} and graph neural networks (\textbf{GNNs}). The motivation was come from the facts that MLNs are computationally intractable from an NP-complete problem and the data-driven GNNs are unable to leverage the domain knowledge. {Fig. \ref{Zhang20}} shows the working principle of ExpressGNN, in which knowledge graph (\textbf{KG}) is a knowledge base representing a collection of interlinked descriptions of entities. ExpressGNN scales MLN inference to KG problems and applies GNN for variational inference. With the GNN controllable embeddings, ExpressGNN were able to achieves more efficiency than MLNs in logical reasoning as well as a balance between expressiveness and simplicity of the model. In the numerical investigations, they demonstrated that ExpressGNN was able to fulfill zero-shot learning (\textbf{ZSL}).  
{Fig. \ref{Socher}} illustrates an example \cite{Socherg:13:NIPS} about ZLS, which is a learning to recognize unseen visual classes with some semantic information (or GCs) given. The term ``{\it zero-shot}'' means no instance to be given in the training dataset. ZLS is a kind of GCL because semantic information and/or unseen classes may be ill-defined or incompletely described in a form of GCs. Semantic gap may be involved for GMs, both in encoding and decoding of GCs, such as discovering unseen classes. 


\begin{figure}
	\centering
	\includegraphics[width=0.920\linewidth]{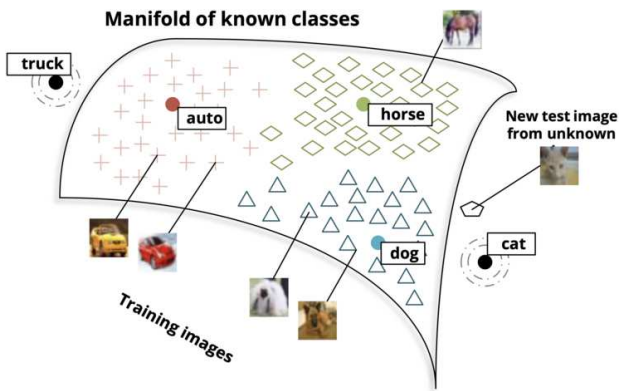}\\
	\caption{Example of zero-shot learning \cite{Socherg:13:NIPS} for classifying unseen classes in the training data, where the unseen classes are truck and cat with some semantic information given.
	}
	\label{Socher}
\end{figure}

Another example is about studying co-evolving financial time-serious problems from GMs. In \cite{Bai:20:TNNLS}, Bai {et al.} proposed a tool, called EDTWK (Entropic Dynamic Time Warping Kernels), for time-varying financial networks. They computed the commute time matrix on each of the network structures for satisfying a GC as ``{\it the financial crises are usually caused by a set of the most mutually correlated stocks while having less uncertainty} \cite{Haubrich:13:Book, Bai:20:TNNLS}''. Based on the commute time matrix, a dominant correlated stock set was identified for processing.  They conducted numerical investigations from the New York Stock Exchange (\textbf{NYSE}) dataset. For the given data, EDTWK was formed as a family of time-varying financial network with a fixed number of 347 vertices from 347 stocks and varying edge weights for the 5976 trading days. They used EDTWK to classify the time-varying financial networks into corresponding stages of each financial crisis period. Five sets of the crisis periods were studied, that is, {\it Black Monday}, {\it Dot-com Bubble}, {\it 1997 Asia Financial Crisis}, {\it Newcentury Financial Bankruptcy}, and {\it Lehman Crisis}. EDTWK was able to detect them from the given data and to characterize different stages in time-varying financial network evolutions. Some crisis events were shown in 3-D embedding, such as the two plots from two events in {Fig. \ref{Bai20}}. One can see that \textbf{the main challenge is still about how to discover and abstract GCs from the model and its visualization, either for modelers or for machines}. 

The examples given above indicate that, in AI modeling, it is a promising direction of merging other models with GMs, or utilizing graphic representations as much as possible. The studies in natural language processing (\textbf{NLP}) \cite{Mikolov:13:arXiv, Le:14:arXiv, Peng:18:WWW} have demonstrated that GMs can process natural language and have achieved much better performance than the conventional NLP models. Due to a transparent feature of GMs \cite{Koller:09:Book}, more studies have been reported, such as on GNNs \cite{Zhou:18:arXiv, Zhang:20:TKDE, Wu:20:TNNLS} and references therein. 

\begin{figure}
  \centering
  \begin{tabular}{@{}c@{}}
    \includegraphics[width=.7\linewidth,height=120pt]{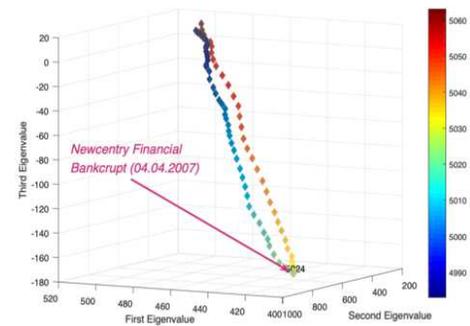} \\[\abovecaptionskip]
    \small (a) 
  \end{tabular}

  \vspace{\floatsep}

  \begin{tabular}{@{}c@{}}
    \includegraphics[width=.7\linewidth,height=120pt]{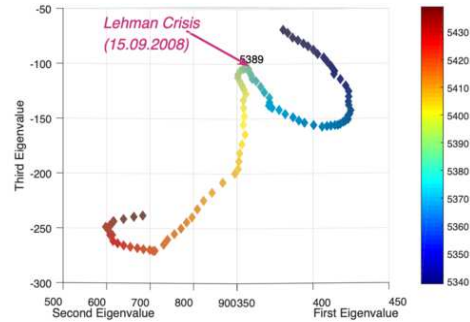} \\[\abovecaptionskip]
    \small (b) 
  \end{tabular}

\caption{Time-varying Trajectory in 3-D embedding for (a) {\it Newcentury Financial Bankruptcy} (4th April 2007) and (b) {\it Lehman Brothers Bankruptcy} (15th September 2008), where the data in 90 trading days (i.e., 90 points) were taken around each of the two events. \cite{Bai:20:TNNLS}.}\label{Bai20}
\end{figure}

\section{Summary and final remarks}
\label{section5}

In this paper, we presented a comprehensive review of GCs as a novel mathematical problem in AI modeling. Comparisons between CCs and GCs were given from several aspects, such as problem domains, representation forms, related tasks, etc. in Table I. One can observe that GCs provide a much larger space over CCs in the future AI studies. We introduced the history of GCs and considered the idea behind Zadeh \cite{Zadeh:96:TFS} on GCs quite stimulating for us to go further. Various methods were provided along a hierarchical way in Fig. \ref{Qu11}, so that one can get a direct knowledge about working format behind every method. Although most researchers have not applied the notion of GCs, they originally encountered problems of GCs, rather than CCs. The given examples demonstrated GCs to be a general problem in constructions of AI machines, and exhibited extra challenges. One may question about the ``{\it new mathematical problem}" proposed in  the paper, because it is significantly different with the conventional descriptions in mathematics. The history that Shannon \cite{Shannon:48:BSTJ} developed a mathematical theory of communication from seeking new problems inspired the authors to identify the new mathematical problem as a starting point for a mathematical theory of AI. However, defining new mathematical problems of AI will be far more complicated than the conventions by covering a broad spectrum of disciplines. This paper attempted
to support that \textbf{GCs should be a necessary set in learning target selection \cite{Hu:15:CIAC} towards a mathematical theory of AI. GCs will 
define every learning approaches and their outcomes fundamentally}. 

In the paper, we also presented our perspectives about future AI machines in related to GCs. We advocated the notion of {\it big knowledge} and redefined XAI for pursuing {\it deep knowledge}. GCs will provide a critical solution to achieving XAI. \textbf{We expect that AI machines will advance themselves in terms of explicit knowledge via increased GCs for its {big} and {deep} knowledge goals.} However, increased GCs will bring more theoretical challenges, such as:
\begin{itemize}
	\item Theoretical Challenge 1: What kinds of GCs are intrinsically \textit{undecidable} \cite{Cooper:17:Book}, or \textit{unknowable} \cite{Geer:19:TSAP}? 
\end{itemize}
The challenge itself falls within XAI and implies importance in seeking a \textit{theoretical boundary of AI}. An excellent example is from Turing's study on the halting problem for Turing machines \cite{Turing:37:LMS}. This theoretical study suggests that we need to avoid the efforts of generating a \textit{perpetual motion machine} in AI applications. For the concerns or studies like ``\textit{Can AI takeover humans?}" or ``\textit{superhuman intelligence}'', we suggest that they should be finally answered in explanations of mathematics, rather than only in descriptions of linguistic arguments. 

For the final remarks of the paper, we cite one saying from Confucius and present brief discussions on it below:

\hspace{0.5cm} ``{\it \textbf{Learning without thinking results in} 

\hspace{3.3cm} \textbf{confusion (knowledge). }

\hspace{0.6cm} \textbf{Thinking without learning ends in }

\hspace{3.2cm} \textbf{empty (new knowledge).}}''
 
\hspace{4cm} by Confucius (551–479 BCE)
When Confucius suggested the saying for his students,  we take it by adding key words ``{\it knowledge}" and ``{\it new knowledge}" from AI perspective. The saying above exactly suggests that human-level AI should have both procedures for their developments, i.e., {\it bottom-up learning} and {\it top-down thinking}, where GCs will play a key role in mathematical modeling at both input and output levels for future AI machines.

\appendix[GCs for Fig. 3]
\begin{figure}
	\centering
	\includegraphics[scale=0.2]{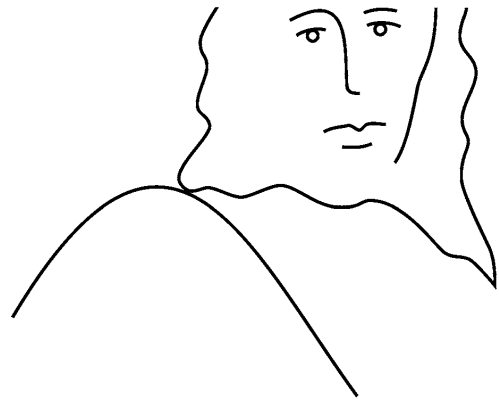}\\. 
	\caption{The approximation \cite{Gordon:04:Book}, or GCs, of the image in Fig. \ref{figPorter}.
	}
	\label{fig_A}
\end{figure}


\ifCLASSOPTIONcompsoc
\else
\fi


\ifCLASSOPTIONcaptionsoff
  \newpage
\fi



\bibliographystyle{IEEEtran}
\bibliography{IEEEabrv.bib,references.bib}
\vspace{-40pt}
\end{document}